\PassOptionsToPackage{table,xcdraw}{xcolor}

\documentclass{article} % For LaTeX2e

\usepackage{iclr2023_conference,times}

\usepackage{amsmath,amsfonts,bm}

\def\eqref#1{equation~\ref{#1}}
\def\1{\bm{1}}

\DeclareMathAlphabet{\mathsfit}{\encodingdefault}{\sfdefault}{m}{sl}
\SetMathAlphabet{\mathsfit}{bold}{\encodingdefault}{\sfdefault}{bx}{n}

\usepackage{subfigure} % 子图包
\usepackage[utf8]{inputenc} % allow utf-8 input
\usepackage[T1]{fontenc}    % use 8-bit T1 fonts
\usepackage{hyperref}       % hyperlinks
\usepackage{url}            % simple URL typesetting
\usepackage{booktabs}       % professional-quality tables
\usepackage{amsfonts}       % blackboard math symbols
\usepackage{nicefrac}       % compact symbols for 1/2, etc.
\usepackage{microtype}      % microtypography
\usepackage{appendix}
\usepackage{tabularx}
\usepackage{wrapfig,lipsum,booktabs}
\usepackage{graphicx}
\usepackage{floatrow}
\newfloatcommand{capbtabbox}{table}[][\FBwidth]

\newcommand{\nomethod}{19}

\title{An Empirical Study on the Efficacy of Deep Active Learning for { Image Classification}}

\author{Yu Li\textsuperscript{1}\footnotemark[1] \quad Muxi Chen\textsuperscript{2}\footnotemark[1] \quad Yannan Liu \textsuperscript{3} \quad Daojing He \textsuperscript{1} \quad  \textbf{Qiang Xu\textsuperscript{2}} \\
\textsuperscript{1}Harbin Institute of Technology (ShenZhen) \ \textsuperscript{2} The Chinese University of Hong Kong \ \textsuperscript{3}ByteDance
\\
\texttt{li.yu@hit.edu.cn} \quad \texttt{\{mxchen21, qxu\}@cse.cuhk.edu.hk} \quad \\ \texttt{{liuyannan.space}@bytedance.com}  \quad  \texttt{{hedaojinghit}@163.com} 
}

\iclrfinalcopy % Uncomment for camera-ready version, but NOT for submission.

\begin{document}

\maketitle
\renewcommand{\thefootnote}{\fnsymbol{footnote}}
\footnotetext[1]{{Equal Contribution}}

\begin{abstract}
% Deep Active Learning (DAL) has been advocated as a promising method to reduce labeling cost in supervised learning. However, existing evaluations of DAL methods are based on different settings, and their results are controversial. To tackle this issue, this paper comprehensively evaluates \nomethod \ existing DAL methods in a uniform setting, including traditional fully-\underline{s}upervised \underline{a}ctive \underline{l}earning (SAL) strategies and emerging \underline{s}emi-\underline{s}upervised \underline{a}ctive \underline{l}earning (SSAL) techniques. We have several non-trivial findings. First, most SAL methods cannot achieve higher accuracy than random selection. 
% % Among the few strategies that are better than random selection, there is no clear winner, i.e., no SAL method can consistently outperforms other solutions. 
% Second, semi-supervised training brings significant performance improvement compared to pure SAL methods. Third, performing data selection in the SSAL setting can achieve significant and consistent performance improvement, especially with abundant unlabeled data. Our findings produce the following guidance for practitioners: one should
% (i) apply SSAL as early as possible and (ii) collect more unlabeled data whenever possible, for better model performance. We will release our code upon acceptance.

Deep Active Learning (DAL) has been advocated as a promising method to reduce labeling costs in supervised learning. However, existing evaluations of DAL methods are based on different settings, and their results are controversial. To tackle this issue, this paper comprehensively evaluates \nomethod \ existing DAL methods in a uniform setting, including traditional fully-\underline{s}upervised \underline{a}ctive \underline{l}earning (SAL) strategies and emerging \underline{s}emi-\underline{s}upervised \underline{a}ctive \underline{l}earning (SSAL) techniques. We have several non-trivial findings. First, most SAL methods cannot achieve higher accuracy than random selection. Second, semi-supervised training brings significant performance improvement compared to pure SAL methods. Third, performing data selection in the SSAL setting can achieve a significant and consistent performance improvement, especially with abundant unlabeled data. Our findings produce the following guidance for practitioners: one should
(i) {  apply SSAL early} and (ii) collect more unlabeled data whenever possible, for better model performance. We will release our code upon acceptance.

\end{abstract}

\section{Introduction}
\label{sec:intro}
\begin{figure}[t]
    \centering
    \includegraphics[width=0.95\linewidth]{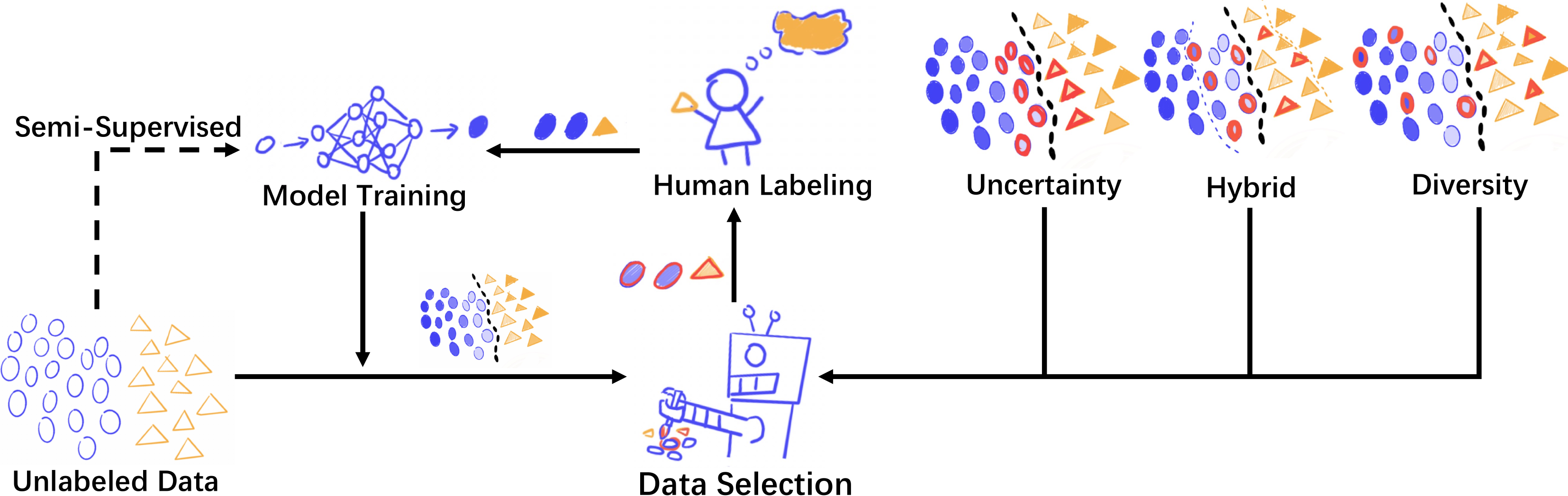}
    \caption{The pool-based deep active learning process. In each iteration, we first train the task model. Based on the trained model and the available unlabeled data, we select a subset of the unlabeled data for labeling (marked as red circle or triangle). The process is iterated until certain model accuracy is achieved or the labeling budget is used up. Existing methods mainly focus on the data selection strategy or the model training strategy.}
    \label{fig:dal-methods}
\end{figure}

Training a well-performed Deep Neural Network (DNN) generally requires a substantial amount of labeled data. However, data collection and labeling can be quite costly, especially for those tasks that require expert knowledge (e.g., medical image analysis \citep{HoiJZL06} and malware detection \citep{NissimCMSEBE14}). Deep Active Learning (DAL) has thus long been advocated to mitigate this issue, wherein we proactively select and label the most informative training samples. That is, given a pool of unlabeled data, DAL iteratively performs data selection and training until the given labeling budget is reached, as shown in Figure \ref{fig:dal-methods}. 

%process consists of many iterations, and each iteration contains two stages: the data selection stage and the training stage. In the data selection stage, a DAL technique selects a batch of informative samples from a pool of unlabeled samples to label, and the training stage trains the model with these labeled data. 

Various DAL techniques are proposed in the literature { for the image classification task}. Most of them are fully-supervised ones (SAL) and aim for a better data selection strategy\footnote{Among the \nomethod \ investigated methods, 14 of them are fully-supervised ones}. The SAL strategies can be roughly grouped into three categories: 1) \textit{model-based selection}; 2) \textit{data distribution-based selection}; and 3) \textit{hybrid selection}. Model-based selection prefers annotating data that are most uncertain under the task model~%, e.g., using model outputs to measure the uncertainty of unlabeled samples 
\citep{gal2017deep, Beluch2018ThePO}. Data distribution-based methods select data according to their density or diversity \citep{iclrSenerS18, iccvSinhaED19}. Hybrid methods consider task model information and data distribution when selecting \citep{iclrAshZK0A20}. 

% Besides the selection strategy, some recent works improve the training stage by leveraging the unlabeled data to enhance the task model's feature extraction ability. 
By applying pseudo-labels~\citep{2019Pseudo} to the unlabeled data or consistency regularization~\citep{mixmatch}, semi-supervised learning (SSL) can improve the model performance substantially. Consequently, it is attractive to apply active learning on top of SSL techniques, referred to as semi-supervised active learning (SSAL).
%Besides the SAL strategy, some recent works improve model performance by directly leveraging the unlabeled data, i.e., semi-supervised learning (SSL) (e.g.,~\citep{mixmatch}). 
% Combining SSL and SAL seems very natural to increase the label efficiency.
%Among these semi-supervised active learning (SSAL) methods, 
\citet{MMA} incorporate the well-known SSL method MixMatch \citep{mixmatch} during training. 
\citet{Consistency} augment the unlabeled samples and enforce the model to have consistent predictions on the unlabeled samples and their corresponding augmentations. 
Also, \citet{Consistency} develop a data selection strategy for the SSL-based method, i.e., selecting samples with inconsistent predictions. Besides, WAAL formulates the DAL as a distribution matching problem and trains the task model with an additional loss evaluated from the distributional difference between labeled and unlabeled data \citep{shui2020deep}.

Despite the effectiveness of existing methods, the results from previous works often contradict each other. 
For example, CoreSet~\citep{iclrSenerS18} and the DBAL method~\citep{gal2017deep} are shown to perform worse than the random selection (RS) in \citep{Beluch2018ThePO} and \citep{iccvSinhaED19}, respectively.
Such result inconsistency is often due to the different experimental settings in evaluation (see Appendix Table \ref{table:hyperparameters}).
Several empirical studies are therefore proposed to address this problem \citep{DBLP:journals/corr/abs-2106-15324, DBLP:journals/corr/abs-2002-09564}. 
However, again, they have controversial observations. \citet{DBLP:journals/corr/abs-2106-15324} claim that there is little or no benefits for different DAL methods over RS, while \citet{DBLP:journals/corr/abs-2002-09564} show that DAL methods are much better than RS.

% \textbf{Are DAL methods really better than random selection (RS) ? In what circumstances the DAL method is better than RS? What guidance we can provide to the DAL community through this empirical study?}

% These inconsistencies motivate us to study the effectiveness of the state-of-the-art DAL methods.
% To answer these questions, we propose to unify the experimental settings and conduct a thorough empirical study on the effectiveness of DAL techniques.
These inconsistencies motivate us to unify the experimental settings and conduct a thorough empirical study on the effectiveness of DAL techniques.
Our contributions are summarized as follows:

% In this paper, we implement and evaluate the performance and cost of \nomethod \ existing state-of-the-art DAL techniques. The solutions been evaluated ranging from the ones focusing on data selection to the ones focusing on leveraging the unlabeled data during training. 
% Second, we study the robustness of these techniques by varying the related hyper-parameters such as the start budget size and the budget for each iteration. 

% Last but not least, we observe a strong correlation between the semi-supervised active learning and the energy-based model, and found two underlying reasons which produces a better model: the model entropy and model smoothness. 

\begin{itemize}
    % \item We re-implement and perform extensive evaluations of \nomethod \ deep active learning methods on various popular image classification tasks, including MNIST, CIFAR-10, and GTSRB. To the best of our knowledge, our evaluation is the most comprehensive one, which not only includes most state-of-the-art solutions but also conducts the first empirical study that incorporates various SSAL methods.
    \item We re-implement and perform extensive evaluations of \nomethod \ deep active learning methods on several popular image classification tasks, including MNIST, CIFAR-10, and GTSRB. To the best of our knowledge, our evaluation is the most comprehensive one, which not only includes most state-of-the-art SAL solutions but also incorporates various SSAL methods.

    \item Through extensive experiments, we conclude that SSAL techniques are preferred. Traditional SAL methods can hardly beat random selection, and there is no SAL method that can consistently outperform others. In contrast, SSAL methods not only easily outperform all the SAL methods by a large margin. More importantly, active sample selection plays an important role in SSAL methods, which can achieve significant and consistent performance improvements compared to random selection.
    
    % \item We conduct an in-depth analysis of SSAL methods and provide two guidance to the practitioners. First, one should conduct SSAL as early as possible. Second, one should seek more unlabeled data whenever possible to achieve better performance.
    \item We conduct an in-depth analysis of SSAL methods and provide two guidance to the practitioners. { First, one should conduct SSAL early}. Second, one should seek more unlabeled data whenever possible to achieve better performance.

\end{itemize}

% This paper studies the effectiveness of existing deep active learning methods. Our study is the most comprehensive one which cover \nonumber \ DAL methods including the SAL and SSAL ones. Through extensive experiments, we observe that the emerging SSAL techniques provide promising direction for the DAL filed. Specifically, the traditional SAL methods can hardly beat random selection, and there is no SAL method can consistently outperform others. However, the SSLAL methods can easily outperform random selection and we find that sample selection can bring huge performance improvement in the SSLAL cycle. In this paper, we also conduct in-depth analysis on the use-case of the SSAL methods and give two recommendations to the practitioners. First, one should conduct SSAL as early as possible and seek for more unlabeled data whenever possible to achieve better performance.

% These observations prove the effectiveness of DAL techniques and show the potential direction of DAL. That is, to have a better model, one should focus on reducing the model entropy and increase the model smoothnes. We leave this as our future work.

The rest of the paper is organized as follows. Section \ref{sec:related} presents the related works. Section \ref{sec:empiricalsetup} illustrates the experimental setup. We present our empirical study on the performance in Section \ref{sec:empiricalresults}. Further studies on SSAL methods are presented in Section \ref{sec:sslal}. Section \ref{sec:conclusion} concludes this paper.

% \section{Problem Formulation and Related Works}
\section{Related Works}
\label{sec:related}

% In this section, we provide a formal definition of the deep active learning (DAL) problem. Then, we introduce related works towards solving the DAL problem.

% \subsection{Problem Formulation}
% Given a pool of unlabeled training data $\mathcal{X} = \{x_i\}_{i=1}^{|X|}$ and a labeling budget $b$, where $\mathcal{X}$ donates the input space. DAL aims to select $b$ samples from $\mathcal{X}$ to label so that the classification accuracy of model $\mathbf{\theta}$ trained on the selected samples is maximized. 

% DNNs training requires a vast quantities of annotated data. However , annotating data is costly and time consuming. With the goal of making deep learning more label-efficient, DAL find effective ways to choose data points that maximizes the accuracy of DNNs to label, from a pool of unlabeled data points.

In this section, we introduce existing DAL methods and empirical studies on these DAL methods.
As shown in Figure \ref{fig:dal-methods}, existing DAL works can be roughly grouped into fully-supervised active learning (SAL) and semi-supervised active learning (SSAL), depending on whether they use unlabeled data to train the task model.

\subsection{Fully-supervised active learning (SAL)}

We categorize the SAL strategies into three classes: uncertainty-based selection, diversity/ representativeness-based selection, and hybrid selection, which is the combination of the above two methods.

\noindent
\textbf{Uncertainty-based selection.} 
The uncertainty-based approaches estimate the uncertainty for each unlabeled sample based on the task model's output and select uncertain data points for labeling. Various methods have been developed to estimate uncertainties.
Least Confidence, Entropy Sampling, and Margin Sampling directly use the model output for an input image for sample uncertainty estimation \citep{DBLP:conf/ijcnn/WangS14}. Specifically, Least Confidence calculates the uncertainty of a sample as $1 - largest \ output \ probability$, Entropy Sampling uses the output entropy to represent the uncertainty, and Margin Sampling uses the difference between the largest and the second largest output probability as the uncertainty of a sample.
BALD Dropout runs the task model multiple times for an unlabeled sample with a certain dropout rate. Data uncertainty is defined by the difference between the mean entropy of multiple probability predictions and the entropy of the mean probability prediction \citep{DBLP:conf/icml/GalIG17}. 
% Some methods leverage additional models to measure the uncertainty of a sample. 
Learning Loss trains an additional module to predict the loss of the original model and regards the loss as sample uncertainty \citep{DBLP:conf/cvpr/YooK19}. 
Uncertain GCN uses a graph convolution network (GCN) to select samples that are unlike the labeled samples \citep{DBLP:conf/cvpr/CaramalauBK21}.
VAAL trains a variational auto-encoder (VAE) and a discriminator to distinguish labeled data and unlabeled data through adversarial learning \citep{iccvSinhaED19}. They regard the probability of belonging to the unlabeled dataset as the uncertainty of a sample.

\noindent
\textbf{Diversity/Representativeness-based selection.}
Diversity/Representativeness-based methods aim at finding a diverse or representative data set from the unlabeled data pool such that a model learned over the subset is competitive over the whole dataset. CoreSet maps the high-dimensional data to low-dimensional feature space and based on which they select the $k$-center points \citep{iclrSenerS18}. KMeans clusters data points into multiple clusters based on their features and selects samples that are closest to the center of each cluster \citep{Zhdanov2019DiverseMA}. CoreGCN is a variant of uncertainGCN which applies the CoreSet algorithm on the feature learned by the GCN \citep{DBLP:conf/cvpr/CaramalauBK21}.

\noindent
\textbf{Hybrid selection.}
In practice, DAL methods typically acquire a batch of data at once. However, on the one hand, samples selected by the uncertainty-based method may contain redundant information. On the other hand, selecting samples without considering the task model's feedback can select samples that the model already accurately infers, thus providing little information to the model. To address this challenge, some methods propose to combine the uncertainty and diversity methods. 
% \begin{itemize}
ActiveLearningByLearning (ALBL) proposes to blend different AL strategies and select the best one by estimating their contribution at each iteration \citep{DBLP:conf/wacv/0007T20}.
BadgeSampling leverages the estimated gradient of the parameters in the last layer for selection. The selected samples have high gradient magnitude and diverse gradient directions \citep{iclrAshZK0A20}.
ClusterMargin clusters the unlabeled data and uses a round-robin strategy to select each cluster's most uncertain data \citep{citovsky2021batch}. 
MCADL considers multiple criteria simultaneously and adaptively fuses the result of uncertainty-based selection and density and similarity-based selection \citep{DBLP:journals/kbs/YuanHXCGN19}.
% \end{itemize}
% For example, Adaptive active learning for image classification \citep{} consider both uncertainty defined by conditional entropy, and information density. 
% Badge proposes that if the label of a data points induces a large gradient of the loss with respect to the model's parameter, the model is uncertain on this data point. And to offset the diversity of data, it selects a set of samples that have both high gradient magnitude and diverse gradient directions with respect to parameters in the output layer. Cluster Margin Sampling \citep{} use a two steps process to select data. It first cluster data into some clusters, then repeatedly selects the most uncertain data points from each cluster, where uncertainty is defined by the difference between the first and second largest probability of classes in classification. 

\subsection{Semi-supervised Active Learning (SSAL)}
% \noindent
% \textbf{Training.}

Another group of works that address the annotation-efficient learning problem is semi-supervised learning (SSL). The main idea of SSL is to mine the useful information in the unlabeled data for model training. One way is to assign each unlabeled sample a pseudo-label and then use these pseudo-labeled samples together with labeled samples for training. Another way is to use consistency regularization evaluated on the unlabeled data for better model training.

Recently, there have been  few attempts to combine SAL and SSL to form the semi-supervised active learning methods (SSAL).
\cite{MMA} incorporates one of the popular SSL techniques MixMatch, into the SAL training pipeline, and the model accuracy is greatly improved. They use two sample selection algorithms, and we denote them as SSLDiff2AugDirect and SSLDiff2AugKmeans:
\begin{itemize}
    \item \textbf{SSLDiff2AugDirect}. The first step of this method is to calculate the average prediction for each sample and its augmentation. Based on the averaged prediction, the authors use the difference between the largest and the second largest probability as the uncertainty for sample selection.
    \item \textbf{SSLDiff2AugKmeans}. They use the k-means algorithm to cluster all unlabeled samples. Then, they select the top uncertain samples from each cluster. The uncertainty is calculated as the SSLDiff2AugDirect method.
\end{itemize}
% At first, they calculate the average prediction on the sample and their augmentation. For SSLDiff2AugDirect, they use the difference between the averaged largest and the second largest probability for sample selection. For SSLDiff2AugKmeans, they use KMeans to cluster the samples based on the averaged prediction and choose the most uncertain samples in each cluster. 
Another SSAL method (we denote it as \textbf{SSLConsistency}) uses consistency regularization to improve the training process. Specifically, for each unlabeled sample, the authors enforce the model to produce consistent predictions on the sample and its augmented version. SSLConsistency also proposes a sample selection mechanism based on this consistency and selects the least consistent ones to label \citep{Consistency}.  We also include the WAAL method as an SSAL because it uses unlabeled data to train the task model. Specifically, WAAL enhances the task model's feature extractor by letting it to differentiate the labeled and unlabeled empirical distributions based on the Wasserstein distance. It selects unlabeled samples that are different from the current labeled data.

{ 
Please note that we also include \textbf{SSLRandom} as a baseline to compare the effectiveness of sample selection in the SSAL setting. SSLRandom selects samples randomly in the sample selection step and uses a semi-supervised training method (e.g., MixMatch) in the model training step.}
% \noindent
% \textbf{Data augmentation.}
% \begin{table}[t]
% \centering
% \caption{Comparison of empirical studies of DAL methods.}
% \resizebox{0.9\linewidth}{!}{  
% \label{table:empiricals}
% \begin{tabular}{c|cccccc}
% \toprule
% Paper & Publication Year & Methods & SSL & Performance & Robustness & Cost \\ \midrule
% Ours & 2022 & 19 & Yes & Yes & Yes & Yes \\  
% \cite{DBLP:journals/corr/abs-2106-15324} & 2021 & 8 & No & Yes & Yes & Yes \\  
% \cite{DBLP:conf/kbse/Hu0CXMPT21} & 2021 & 12 & No & Yes & Yes & No \\  
% \cite{DBLP:journals/corr/abs-2002-09564} & 2020 & 6 & No & Yes & Yes & No \\
% \cite{DBLP:journals/corr/abs-1912-05361} & 2019 & 6 & Yes & Yes & No & No\\ \bottomrule
% \end{tabular}
% }
% \end{table}

\begin{table}[t]
\centering
\caption{{ Comparison of empirical studies of DAL methods. We report the number of total methods (SAL+SSAL) studied in each paper, the number of SSAL methods, and whether they study the robustness and cost of DAL methods. }}
\resizebox{0.8\linewidth}{!}{  
\label{table:empiricals}
\begin{tabular}{c|cccccc}
\toprule
Paper & Total methods & SSAL & Performance & Robustness & Cost \\ \midrule
Ours  & 19 & 5 & Yes & Yes & Yes \\  
\cite{DBLP:journals/corr/abs-2106-15324} & 8 & 0 & Yes & Yes & Yes \\  
\cite{DBLP:conf/kbse/Hu0CXMPT21}         & 12 & 0 & Yes & Yes & No \\  
\cite{DBLP:journals/corr/abs-2002-09564} & 6 & 0 & Yes & Yes & No \\
\cite{DBLP:journals/corr/abs-1912-05361} & 10 & 5 & Yes & No & No \\ 
{  \cite{Chan2021OnTM}}  & 6 & 3 & Yes & No & No\\
\bottomrule
\end{tabular}
}
\end{table}

\subsection{Empirical Study of Active Learning Techniques}
\noindent
Several empirical studies on the effectiveness of DAL methods have been proposed in recent years  \citep{DBLP:journals/corr/abs-2106-15324, DBLP:conf/kbse/Hu0CXMPT21, DBLP:journals/corr/abs-2002-09564, DBLP:journals/corr/abs-1912-05361, Chan2021OnTM}. Recent empirical studies either fail to study DAL methods under a variety of experiment settings or ignore a growing line of work that combines semi-supervised learning with active learning, i.e., SSAL methods. 
We summarize their characteristics in Table~\ref{table:empiricals}. 

% Specifically, 4/5 studies do not the cost of each method. As some well-performed methods can come at large cost, providing the cost can provide more information to the practitioners when applying these methods on their application. 

{ 
Among these five works, only two cover SSAL methods \cite{DBLP:journals/corr/abs-1912-05361, Chan2021OnTM}. 
While they proves that SSAL preforms better than SAL, they do not conduct an in-depth study on the SSAL methods to figure out the important factors that influence the performance of the SSAL methods. 
Besides, \cite{Chan2021OnTM} only use three sample selection methods (e.g., CoreSet and VAAL) for validation and ignore the state-of-the-art SSAL methods (e.g., uncertainty-based ones).
}

In contrast, our work unifies the SAL and SSAL into a single evaluation framework. Also, we provide an in-depth analysis of the SSAL methods. This analysis provides practical guidance to the users.
% consistently performs better than SAL. 
% Distil \citep{DBLP:journals/corr/abs-2106-15324} is a current empirical study of active learning techniques. It studies the effect of data augmentation, redundant data, different label set initializations, and the choice of budget size on several active learning techniques in image classification tasks. Fine-tuning versus retraining and time and energy cost are also studied. However, it neglect a growing line of work that combines semi-supervised learning with active learning. 
% \citep{DBLP:conf/kbse/Hu0CXMPT21} studies the performance of active learning strategies under different model regularization, imbalance dataset, and different initial and query budget sizes.
% \citep{DBLP:journals/corr/abs-2002-09564} studies the robustness of active learning strategies under adversarial attacks and model compression.
% \citep{DBLP:journals/corr/abs-1912-05361} is also an empirical study of active learning methods, which addresses comparing the performance of active learning, semi-supervised learning, data augmentation, and semi-supervised active learning approaches under different budget sizes. As it was published in 2019, only six active learning methods are considered.

\section{Experiment Setup}
\label{sec:empiricalsetup}
% \begin{figure}
%     \centering
%     \includegraphics[width=\linewidth]{figs/Strategies.pdf}
%     \caption{Classification of the DAL methods under study.}
%     \label{fig:classification}
% \end{figure}

\textbf{DAL techniques under study.}
We re-implement and reproduce the results for \nomethod   \ DAL methods, including the fully-supervised active learning (SAL) and the semi-supervised active learning (SSAL). For SAL methods, we cover  \textbf{RandomSampling}, \textbf{ALBL}, \textbf{AdversarialBIM}, \textbf{BALDDropout}, \textbf{BadgeSampling}, \textbf{ClusterMargin}, \textbf{CoreSet}, \textbf{KMeansSampling}, \textbf{LearningLoss}, \textbf{LeastConfidence}, \textbf{MCADL}, \textbf{CoreGCN}, \textbf{UncertainGCN}, and \textbf{VAAL}.
For SSAL methods, we cover \textbf{SSLRandom}, \textbf{SSLConsistency}, \textbf{SSLDiff2AugDirect}, \textbf{SSLDiff2AugKmeans}, and \textbf{WAAL}. 
% We regard the \textbf{WAAL} as a SSAL method because it uses the unlabeled data for task learning. 
For a detailed description of each method, please refer to Section \ref{sec:related}.

% \begin{wraptable}{r}{0.7\linewidth}
% \resizebox{0.7\linewidth}{!}{  
% \begin{tabular}{cccc}
% \toprule
% Dataset       & \# Classes & Train/Test     & Acc (\%) \\ \midrule
% MNIST         & 10         & 60,000/10,000  & 97.68    \\
% CIFAR-10      & 10         & 50,000/10,000  & 92.15    \\
% GTSRB         & 43         & 39,209/12,630  & 94.42    \\
% Tiny-ImageNet & 200        & 100,000/10,000 & 55.85   \\ \bottomrule
% \end{tabular}
% }
% \caption{The dataset and the accuracy of the fully trained model.}

% \end{wraptable} 

\noindent
\textbf{Datasets {  and models}.}  In this paper, we use three widely used image classification datasets for evaluation: MNIST \citep{lecun1998gradient}, CIFAR10 \citep{krizhevsky2009learning}, CIFAR100 \citep{krizhevsky2009learning}, GTSRB \citep{Houben-IJCNN-2013}, and TinyImageNet \citep{Le2015TinyIV}. 
MNIST contains 70,000 examples of handwritten digits. The size of each image is 28 × 28 × 1. 
CIFAR10 is composed of ten classes of natural images and has 60,000 images in total. The size of each image is 32 × 32 × 3. 
{ 
CIFAR100 contains 100 classes of natural images and has 60,000 images in total. The size of each image is 32 × 32 × 3. 
}
GTSRB contains 43 classes of traffic sign images. There are 39,209 labeled images for training and 12630 images for testing. Each image has 32 × 32 × 3 pixels. Please note that the GTSRB dataset has imbalanced classes. 
{ 
Tiny-ImageNet contains 200 classes and 110,000 image in total, including 100,000 images for training and 10,000 images for testing. The image size in Tiny-ImageNet is 64 × 64.
}
{  The model used for each dataset and the fully-supervised model accuracy achieved are shown in Table~\ref{table:datasets}. } 

% \textbf{Model.} 

\noindent
\textbf{Training settings.}
We evaluate all methods with a unified setting, which is adopted from \citep{iclrAshZK0A20}. Specifically, for MNIST, we set the start budget (denoted as nStart) as 2\%, the total budget (denoted as nEnd) as 20\%, the step size (denoted as nQuery) as 2\%, and the training epochs as 50. That is to say, we will randomly select 2\% samples from the dataset to initialize the task model. Then, we query 2\% extra unlabeled samples in each iteration for labeling. The process is iterated until the labeled set size reaches 20\%. For the rest of the datasets, we set the start point nStart as 10\%, the total budget nEnd as 70\%, the step size nQuery as 5\%, and the training epochs as 150.
We use SGD as the optimizer since it performs better than Adam \citep{DBLP:journals/corr/abs-2106-15324}. The initial learning rate is set to 0.1 for all datasets.
{  Also, we include two learning rate decay stages. In each stage, we decay the learning rate by 0.1. For MNIST, we set the first and the second decay stage at epoch 20 and 40, respectively. Since the rest datasets needs more epochs to train the model, we set the the decay stages at epoch 80 and 120, respectively. }
According to \citep{DBLP:journals/corr/abs-2106-15324}, resetting the model after each iteration achieve better performance than fine-tuning the model from previous rounds. Hence, we reset the model for each DAL iteration and retrain the model from scratch. 

Please note that we evaluate each method for five runs, and we try to fix the five random seeds so that every method has the same initial accuracy. However, there are some exceptions: WAAL and LearningLoss. This is because WAAL uses unlabeled data for training, and the model learned by learning loss will be affected by the loss model.

\textbf{Platform and Hardware.} We use the Pytorch framework and two Titan V GPUs for evaluation.

\textbf{Evaluation metrics.} In this paper, we use three metrics to evaluate different DAL methods: \emph{Accuracy}, \emph{Average Performance Improvement (API)}, and \emph{Cost}. The accuracy of a trained model is defined as the ratio between the correctly classified test inputs out of total test inputs.
Due to the large number of methods we study, plotting the accuracy for all methods at each iteration in a single figure is hard for visualization. Hence, we use API, i.e., the average performance of a DAL method over all iterations. API is calculated as follows:

\begin{equation}
    API_{al} = \frac{1}{N} \sum^{N}_{i = 1} (acc^{al}_i - acc^{random}_i),
\end{equation}
where $N$ represents the total iterations, i.e., $N = 1 + int(\frac{nEnd - nStart}{nStep})$. Also, the $acc^{al}_i$ stands for the accuracy achieved by a deep active learning method (e.g., CoreSet) at iteration $i$. The $acc^{random}_i$ stands for the accuracy achieved by {RandomSampling} at iteration $i$.

Besides, we also report the cost of each deep active learning method as the running time in seconds. The cost mainly contains two parts: the data selection part and the training part. The calculation of the cost is as follows:
\begin{equation}
    Cost_{al} = \frac{1}{N} \sum^{N}_{i = 1} (selection\_cost^{al}_i + training\_cost^{al}_i),
\end{equation}
where $N$ represents the total iterations. The $selection\_cost^{al}_i$ and $training\_cost^{al}_i$ represents the selection cost and the training cost for certain active learning method at iteration $i$.
% For better understanding, we normalize the cost with regard to RandomSampling.

% \begin{table}
% \label{table:datasets}
% \caption{The dataset and the fully-supervised model accuracy.}
% \centering
% \resizebox{0.5\linewidth}{!}{  
% \begin{tabular}{cccc}
% \toprule
% Dataset       & \# Classes & Train/Test     & Acc (\%) \\ \midrule
% MNIST         & 10         & 60,000/10,000  & 97.68    \\
% CIFAR-10      & 10         & 50,000/10,000  & 92.15    \\
% GTSRB         & 43         & 39,209/12,630  & 99.33    \\
% % Tiny-ImageNet & 200        & 100,000/10,000 & 55.85   \\
% \bottomrule
% \end{tabular}
% }
% \end{table} 

% \begin{table}
% \label{table:datasets}
% \caption{{ The dataset and the fully-supervised model accuracy.}}
% \centering
% \resizebox{0.6\linewidth}{!}{  
% \begin{tabular}{ccccc}
% \toprule
% Dataset       & \# Classes & Train/Test     & Model & Acc (\%) \\ \midrule
% MNIST         & 10         & 60,000/10,000  & LeNet & 97.68    \\
% CIFAR-10      & 10         & 50,000/10,000  & ResNet-18 & 92.15    \\
% GTSRB         & 43         & 39,209/12,630  &ResNet-18  & 99.33    \\
% % Tiny-ImageNet & 200        & 100,000/10,000 & 55.85   \\
% \bottomrule
% \end{tabular}
% }
% \end{table} 

\begin{table}
\label{table:datasets}
\caption{{ The dataset and the fully-supervised model accuracy.}}
\centering
\resizebox{0.68\linewidth}{!}{  
\begin{tabular}{ccccc}
\toprule
Dataset       & \# Classes & Train/Test     & Model & Acc (\%) \\ \midrule
MNIST         & 10         & 60,000/10,000  & LeNet & 97.68    \\
CIFAR-10      & 10         & 50,000/10,000  & ResNet-18 & 92.15    \\
{ CIFAR-100} & 100        & 50,000/10,000 &ResNet-18 & 66.78   \\
{GTSRB}         & 43         & 39,209/12,630  &ResNet-18  & 99.33    \\
{ Tiny-ImageNet} & 200        & 100,000/10,000 &ResNet-18 & 55.81   \\
\bottomrule
\end{tabular}
}
\end{table}

\section{Experimental Results and Analysis}
\label{sec:empiricalresults}
\subsection{Performance comparison}

% \textbf{Performance.} 
We report the performance of each method in Table \ref{table:auc}. We break the results into the SAL part and the SSAL part. 
% The rows in the first part show the performance of the fully-supervised active learning (SAL), and the row in the second part shows the performance of the semi-supervised active learning methods (SSAL). 
For ease of comparison, we report the average accuracy improvement over \textbf{RandomSampling}. { Please note that some clustering-based methods are time-consuming on Tiny-ImageNet. Due to time and resource limitation, we leave it empty.} Besides, we show the detailed accuracy vs. budget curve of these techniques in the Appendix (see Figure \ref{fig:performance} in Appendix).
% Please note that, due to the high training cost associated with Tiny-ImageNet, we only report the results for eight methods.
We have several observations from the results.

\begin{table}[t]
\centering
\caption{The performance and cost comparison of different DAL methods. The performance is calculated as the \textbf{average accuracy improvement (API)} over RandomSampling. We bold the best method from both SAL and SSAL methods. Also, we highlight the rows that consistently outperform the RandomSampling.  \\}

\label{table:auc}
\resizebox{\linewidth}{!}{  
\begin{tabular}{c|ccccc}
\toprule
                                        %  & \multicolumn{5}{c|}{Performance}                           \\ \midrule
Strategy                                                 & MNIST  & CIFAR10 & GTSRB & { CIFAR100} & { Tiny-ImageNet}  \\ \midrule
\multicolumn{6}{c}{The SAL methods} \\ \midrule
\textbf{RandomSampling}                                           &  0.00        &0.00   & 0.00 & 0.00 & 0.00       \\
\textbf{ALBL} \citep{DBLP:conf/wacv/0007T20}   &  -0.15       &1.58   & 0.34  & 2.13 & 0.48      \\
\textbf{AdversarialBIM} \citep{ducoffe2018adversarial}             &  1.44        &0.25   & -0.24  & 0.58 & -6.40              \\
\textbf{BALDDropout} \citep{DBLP:conf/icml/GalIG17}                &  1.18        &1.48   & -0.04  & 1.55 & -0.34      \\
\rowcolor[HTML]{C0C0C0}
\textbf{BadgeSampling} \citep{iclrAshZK0A20}                       &  \textbf{1.84}       &1.89   & 0.27  & 1.72 & -              \\
\rowcolor[HTML]{C0C0C0}
\textbf{ClusterMargin} \citep{citovsky2021batch}           &  1.76        &0.28   & 0.23   & 1.47 & -           \\
\textbf{CoreSet} \citep{iclrSenerS18}                              &  -1.15       &0.95   & 0.26  & 2.07 & 0.57         \\
\textbf{KMeansSampling} \citep{Zhdanov2019DiverseMA}               &  -0.22       &0.17   & 0.20   & -0.27 & -             \\
\textbf{LearningLoss} \citep{DBLP:conf/cvpr/YooK19}                &  -2.32       &-2.38  & -0.94  & -1.31 & 0.05      \\
\rowcolor[HTML]{C0C0C0}
\textbf{LeastConfidence} \citep{DBLP:conf/ijcnn/WangS14}           &  1.40        &\textbf{1.95}   & \textbf{0.34}  & 2.21 & 0.17          \\
\textbf{MCADL} \citep{DBLP:journals/kbs/YuanHXCGN19}               &  0.57        &-0.50   & 0.03   & -1.27 & -1.51                 \\
\textbf{CoreGCN} \citep{DBLP:conf/cvpr/CaramalauBK21}              &  1.21        &1.16   & -8.52  & 1.35 & -5.19              \\
\textbf{UncertainGCN} \citep{DBLP:conf/cvpr/CaramalauBK21}         &  -0.85       &0.14   & -8.77  & -0.07 & -5.37              \\ 
\textbf{VAAL} \citep{iccvSinhaED19}                                &  -0.34       &-0.49  & -24.79  & -2.01 & -0.32       \\ \midrule
\multicolumn{6}{c}{The SSAL methods} \\ \midrule
\rowcolor[HTML]{C0C0C0}
\textbf{WAAL} \citep{shui2020deep}                                 &  3.00        &2.05   & 0.21  & -0.12 & -3.64            \\
\rowcolor[HTML]{C0C0C0}
\textbf{SSLConsistency} \citep{Consistency}                        &  3.39        &\textbf{5.83}   & \textbf{1.19} & 8.099 & 7.65         \\
\rowcolor[HTML]{C0C0C0}
\textbf{SSLDiff2AugDirect} \citep{MMA}                             &  3.49        &5.78   & \textbf{1.19} & 8.988 & -           \\
\rowcolor[HTML]{C0C0C0}
\textbf{SSLDiff2AugKmeans} \citep{MMA}                             &  \textbf{3.50}         &5.74   & 1.17   & 8.913 & 7.68              \\
\rowcolor[HTML]{C0C0C0}
\textbf{SSLRandom}                                                &  1.68        &4.57   & 1.11  & 8.014 & 7.56      \\ \bottomrule
    
\end{tabular}
}
\end{table}

% \textbf{Comparison among different SAL methods.} For selection-based methods, we observe that RandomSampling is a strong baseline. For each dataset, many selection-based methods perform worse than RandomSampling according to Table \ref{table:auc}. To be specific, there are 6/14, 3/14,  6/14, and 2/6 methods perform worse than the RandomSampling for MNIST, CIFAR-10, Tiny-ImageNet, and GTSRB, respectively. Also, most methods perform worse than RandomSampling in at least one dataset. Only BadgeSampling, ClusterMargin, and LeastConfidence perform constantly better than RandomSampling on all studied datasets. 

\textbf{Comparison among different SAL methods.} We observe that RandomSampling is a strong baseline for SAL methods. For each dataset, many SAL methods perform worse than RandomSampling according to Table \ref{table:auc}. To be specific, there are 6/14, 3/14, and 6/14 methods perform worse than RandomSampling for MNIST, CIFAR-10, and GTSRB, respectively. Also, most methods perform worse than RandomSampling in at least one dataset. Only BadgeSampling, ClusterMargin, and LeastConfidence constantly perform better than RandomSampling on all studied datasets. 

Another important observation for SAL methods is that: there is no SAL method that can outperform others on every dataset. For instance, out of the three methods that are better than RandomSelection, BadgeSampling outperforms the other two methods on MNIST, but it performs worse than LeastConfidence on GTSRB and CIFAR-10. More studies on SAL methods can be found in Appendix.
% We hypothesis the underlying reason is related to the dataset properties, and each method has its application scenarios.
% The underlying reason can be related to the dataset properties, and each method has its application scenarios.
% We will illustrate this further in Section \ref{sec:discussion}.

\textbf{Comparison between SAL and SSAL methods.} When comparing SSAL with SAL, we have the following observations. First, in general, SSAL methods outperforms SAL methods for a large margin. On the one hand, the SSLRandom method has a 1.68, 4.57, and 1.11 average accuracy improvement over the RandomSampling on MNIST, CIFAR-10, and GTSRB, respectively. On the other hand, the SSAL improves the best method in SAL by 3.5-1.84=1.66, 5.83-1.95=3.88, 1.19-0.34=0.85 w.r.t. the API on MNIST, CIFAR-10, and GTSRB respectively. 
Second, SSAL methods are more stable than SAL methods. Each SSAL method outperforms the RandomSampling consistently on all datasets. 

This indicates that using unlabeled data to train the task model is useful. The unlabeled data provides extra information for the model to learn so that the ability of the model can be improved. 

% \begin{wrapfigure}{r}{8cm}
%     \centering
%     \includegraphics[width=\linewidth]{figs/SSLhist.pdf}
%     \caption{Caption}
%     \label{fig:ssl_al_bar}
% \end{wrapfigure}

\begin{figure}[t]
    \centering
    \includegraphics[width=0.8\linewidth]{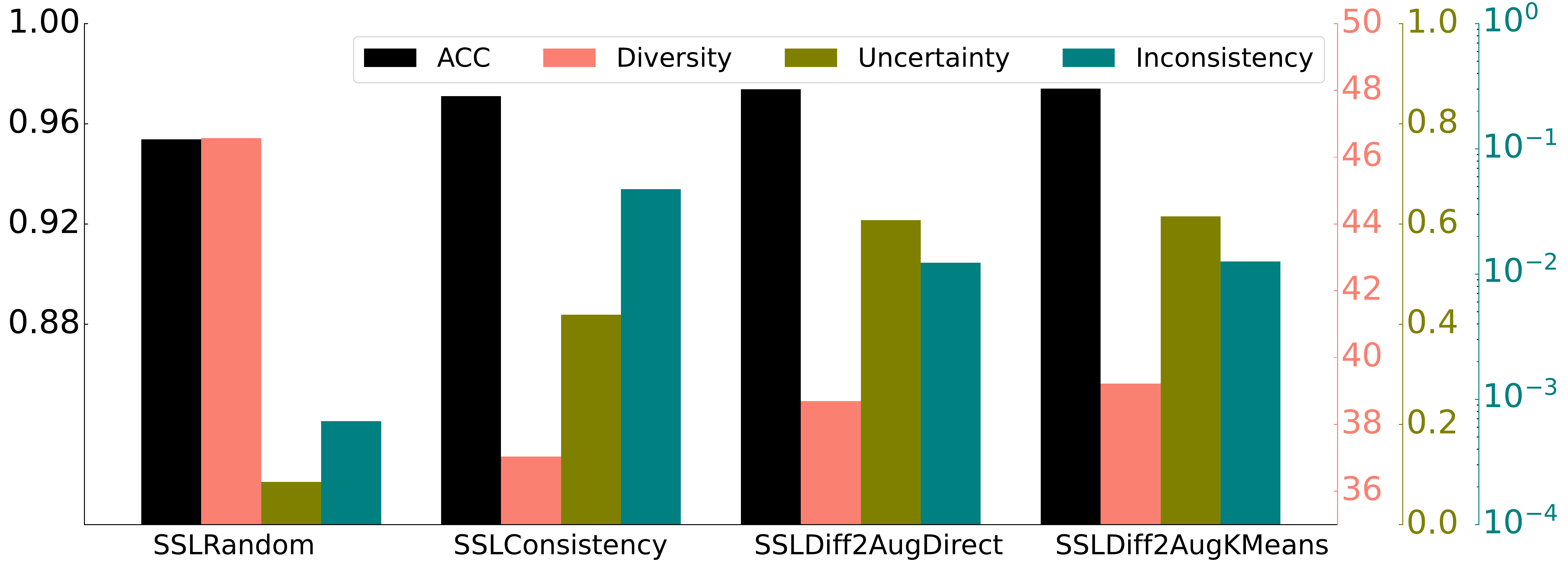}
    \caption{Comparison among different SSAL methods. These four methods use the same SSL method to train the model. This figure is produced using MNIST dataset.}
    \label{fig:ssl_al_bar}
\end{figure}

\begin{figure}[t]
    \centering
    \includegraphics[width=\linewidth]{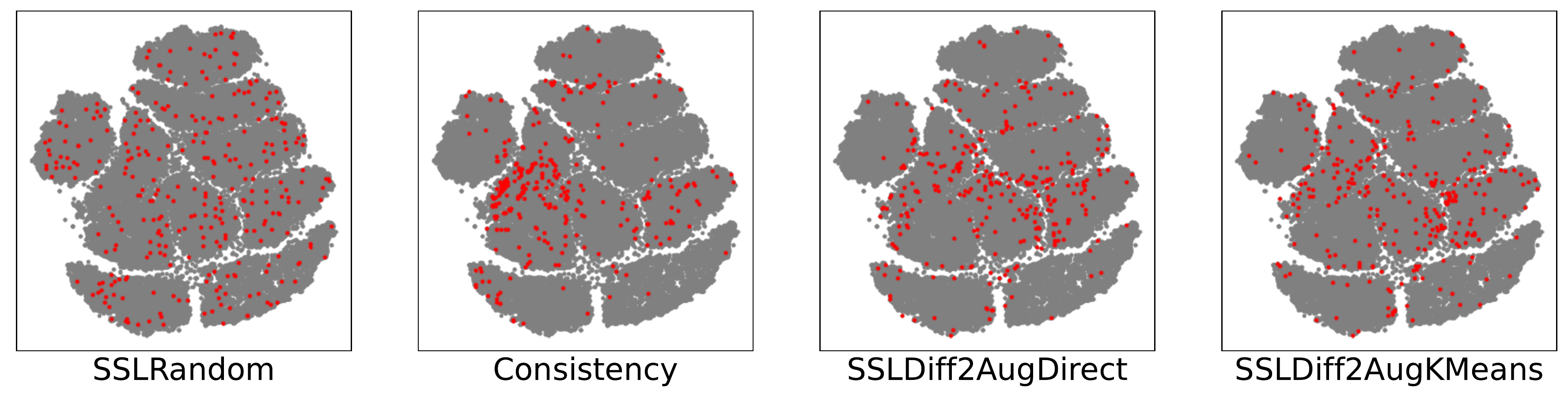}
    \caption{Visualization of selected samples (red points) out of total samples (gray) for different SSAL methods. This figure is produced using the MNIST dataset.}
    \label{fig:ssl_al_tsne}
\end{figure}

\textbf{Comparison among different SSAL methods.} 
% Within the SSL-AL methods, we mainly compare {SSLConsistency}, {SSLDiff2AugDirect}, {SSLDiff2AugKmeans}, and {SSLRandom}. 
% For these methods, we observe that {SSLConsistency}, {SSLDiff2AugDirect}, and {SSLDiff2AugKmeans}, are much better than random selection {SSLRandom}. 
Within the SSAL methods, we observe that {SSLConsistency}, {SSLDiff2AugDirect}, and {SSLDiff2AugKmeans} are much better than {SSLRandom}, while WAAL sometimes performs worse than SSLRandom. 
We also observe that the performance difference between {SSLConsistency}, {SSLDiff2AugDirect}, and {SSLDiff2AugKmeans} are negligibly small.
% Please note that SSLRandom, {SSLConsistency}, {SSLDiff2AugDirect}, and {SSLDiff2AugKmeans} uses the same semi-supervised training method, and their difference can only come from the selection strategies.
% To be specific, SSLRandom selects labeled data randomly. SSLDiff2AugDirect mainly considers the uncertainty of the unlabeled samples, while SSLDiff2AugKmeans considers both the uncertainty and diversity of the selected samples. 

We investigate the reason behind the above observations by studying the property of the selected data with a case study, as shown in Figure \ref{fig:ssl_al_bar} and Figure \ref{fig:ssl_al_tsne}. 
To generate these two figures, we randomly label 2\% samples and train a model using SSL methods. Then, we select and label 1\% extra samples to retrain the model and report the resulting accuracy. 
To simplify the process, we only experiment on SSLRandom, {SSLConsistency}, {SSLDiff2AugDirect}, and {SSLDiff2AugKmeans}. This is because they use similar semi-supervised training methods, and their difference largely comes from the selection strategies.
Figure \ref{fig:ssl_al_bar} shows the diversity, uncertainty, and consistency of the selected samples at certain iteration.
The diversity of the selected samples is calculated as the average distance between selected samples. The uncertainty is calculated as $1-prob$, where $prob$ stands for the highest confidence for one sample. The inconsistency is calculated as the prediction variations between one sample and its augmented version.
Figure \ref{fig:ssl_al_tsne} visualizes the selected samples among all samples. 

{  Figure \ref{fig:ssl_al_tsne} serves as evidence for Figure \ref{fig:ssl_al_bar} to show that the calculation of diversity and uncertainty/consistency is correct. 
As we observe from Figure \ref{fig:ssl_al_bar}, the selected sample by SSLRandom is more diverse than that of other methods, and the distribution of selected samples in Figure \ref{fig:ssl_al_tsne} confirmed this. Also, we observe from Figure \ref{fig:ssl_al_bar} that SSLRandom selects less uncertain samples, and this can be confirmed by Figure \ref{fig:ssl_al_tsne} because samples selected by SSLRandom are far from the decision boundary. 
}

Through this case study, we have the following observations:

% First, we investigate why the performance difference between SSLRandom and the other three methods (e.g., {SSLConsistency}, {SSLDiff2AugDirect}, and {SSLDiff2AugKmeans}) is large. The performance difference shows that the data selection in the SSAL setting brings large benefits. Figure \ref{fig:ssl_al_bar} and Figure \ref{fig:ssl_al_tsne} shows that samples chosen by {SSLRandom} are much more certain and consistent. 
% % These samples may have a high probability of being correctly inferred and hence brings little information for the model to learn. 
% In contrast, the rest three methods focus more on uncertain or inconsistent samples. 
% This indicates that choosing uncertain or inconsistent samples in the SSAL setting can provide more information for the model to learn and bring consistent benefits. 
% Remember that in the SAL setting, most selection methods perform worse than RandomSampling.
% This may be because semi-supervised learning provides the model with better feature extraction ability. Hence, they can estimate the uncertain samples more accurately than in the SAL setting. 

First, we find that choosing uncertain or inconsistent samples in the SSAL setting can improve the model performance consistently and significantly. 
% First, the better feature extraction ablity of SSL mentods leads SSAL methdos out perform SSLRandom and SAL methods significantly.
This is why the performance difference between SSLRandom and the other three methods (e.g., {SSLConsistency}, {SSLDiff2AugDirect}, and {SSLDiff2AugKmeans}) is large. 
To illustrate, Figure \ref{fig:ssl_al_bar} and Figure \ref{fig:ssl_al_tsne} shows that samples chosen by {SSLRandom} are much more certain and consistent. 
In contrast, the rest three methods focus more on uncertain or inconsistent samples. 
Remember that in the SAL setting, most selection methods perform worse than RandomSampling.
This may be because semi-supervised learning provides the model with better feature extraction ability. Hence, they can estimate the uncertain samples more accurately than in the SAL setting. 
% Hence, active sample selection in the SSAL setting is meaningful.

% Second, we investigate why the performance difference between {SSLConsistency}, {SSLDiff2AugDirect}, and {SSLDiff2AugKmeans} are small.  Although {SSLDiff2AugKmeans} considers diversity during sample selection, the performance gain is tiny on MNIST, and there is no performance gain on CIFAR-10 and GTSRB. This result indicates that sample diversity is not as useful as sample uncertainty in the SSAL setting. The underlying reason may be that the learning procedure has already utilized all the unlabeled data and learned the diverse information in these data. Hence, focusing on uncertain samples is more urgent. 

% Second, sample uncertainty is more important than sample diversity during sample selection in SSAL. Since {SSLConsistency}, {SSLDiff2AugDirect}, and {SSLDiff2AugKmeans} all focus on selecting uncertain samples, the performance differences between them are small. To illustrate, although {SSLDiff2AugKmeans} considers diversity during sample selection, the performance gain is tiny on MNIST, and there is no performance gain on CIFAR-10 and GTSRB. This result indicates that sample diversity is not as useful as sample uncertainty in the SSAL setting. The underlying reason may be that the learning procedure has already utilized all the unlabeled data and learned the diverse information in these data. Hence, focusing on uncertain samples is more urgent. 

Second, sample uncertainty is more important than sample diversity during sample selection in SSAL.
{ 
On the one hand, as we can observe from Table \ref{table:auc}, although {SSLDiff2AugKmeans} considers diversity during sample selection, the performance gain is tiny on MNIST, and there is no performance gain on CIFAR-10 and GTSRB. 
In contrast, the uncertainty-based methods (e.g., {SSLConsistency}, {SSLDiff2AugDirect}, and {SSLDiff2AugKmeans}) perform much better than the ones without uncertainty (e.g., SSLRandom). 
% This result indicates that sample diversity is not as useful as sample uncertainty in the SSAL setting.
On the other hand, as we can observe from the case study in Figure \ref{fig:ssl_al_bar}, the sample diversity does not have a clear relationship with the achieved accuracy, but the uncertainty does (higher uncertainty often means higher performance).
We suspect the underlying reason could be that the SSAL learning process already uses diverse information in the unlabeled data. Once the uncertain ones are solved, the diverse information in the rest samples can be utilized naturally. }

Hence, we conclude that active sample selection can significantly improve performance under the SSL setting. That is to say, SSAL is helpful for model performance improvement. Second, we prefer uncertain samples to diverse samples when applying SSAL methods since uncertain samples can bring better results. 
% Choosing uncertain samples can bring better results and is often more cost-friendly than diversity-based selection.
% Therefore, applying active learning in SSL-based methods can improve the performance. In contrast, for non-SSL models, applying active learning methods can hardly have performance gain because they cannot have an accurate estimation of the uncertainties.

% Hence, we conclude that, active learning can have stable improvement when the feature extractor is already well-trained and can accurately estimate the value of the unlabeled data; otherwise, when the estimation is inaccurate, the selected samples can be unpredictable and the result can be worse than random sampling. 

\begin{table}[]
\label{table:cost}
\centering
\resizebox{0.7\linewidth}{!}{  
\begin{tabular}{cccccc}
\toprule
Method & MNIST & CIFAR-10 & GTSRB & CIFAR-100 & Tiny-ImageNet \\ \toprule
SAL    & 3.13  & 2.01     & 1.63  &  2.43   &  3.81             \\ \midrule
SSAL   & 11.95 & 6.03     & 4.05  &  3.89   &  7.26             \\ \bottomrule
\end{tabular}
}
\caption{{ The average cost of SAL and SSAL methods. The cost is normalized using the cost of RandomSampling.}}
\end{table}

\subsection{Cost comparison}
% \textbf{Cost.} 
The cost of SAL and SSAL methods is shown on the Table \ref{table:cost}. We normalize the cost of each method w.r.t. the RandomSampling method. 
% The original time cost for RandomSampling on MNIST, CIFAR-10, GTSRB, and TinyImageNet are 131, 1047, 658, and 1906 seconds, respectively. 
% The original time cost for RandomSampling on MNIST, CIFAR-10, and GTSRB are 131, 1047, and 658 seconds, respectively. 
We have the following observations. First, SSAL methods introduce much more time cost than SAL methods. For example, SSAL methods consume around 4x energy than SAL methods on MNIST. This is because the semi-supervised training will evaluate every unlabeled and labeled sample to calculate the loss. This also means that when the number of unlabeled data increases, the training time will increase. For example, the increase of cost on MNIST (e.g., around 4x) is more than that on GTSRB (e.g. around 3x), around because the number of unlabeled samples in MNIST is more than that in GTSRB.

\begin{figure*}[t]	
    \hspace{0.8cm}
	\subfigure[MNIST] %第一张子图
	{
		\begin{minipage}{5cm}
			\centering          %子图居中
			\includegraphics[scale=0.35]{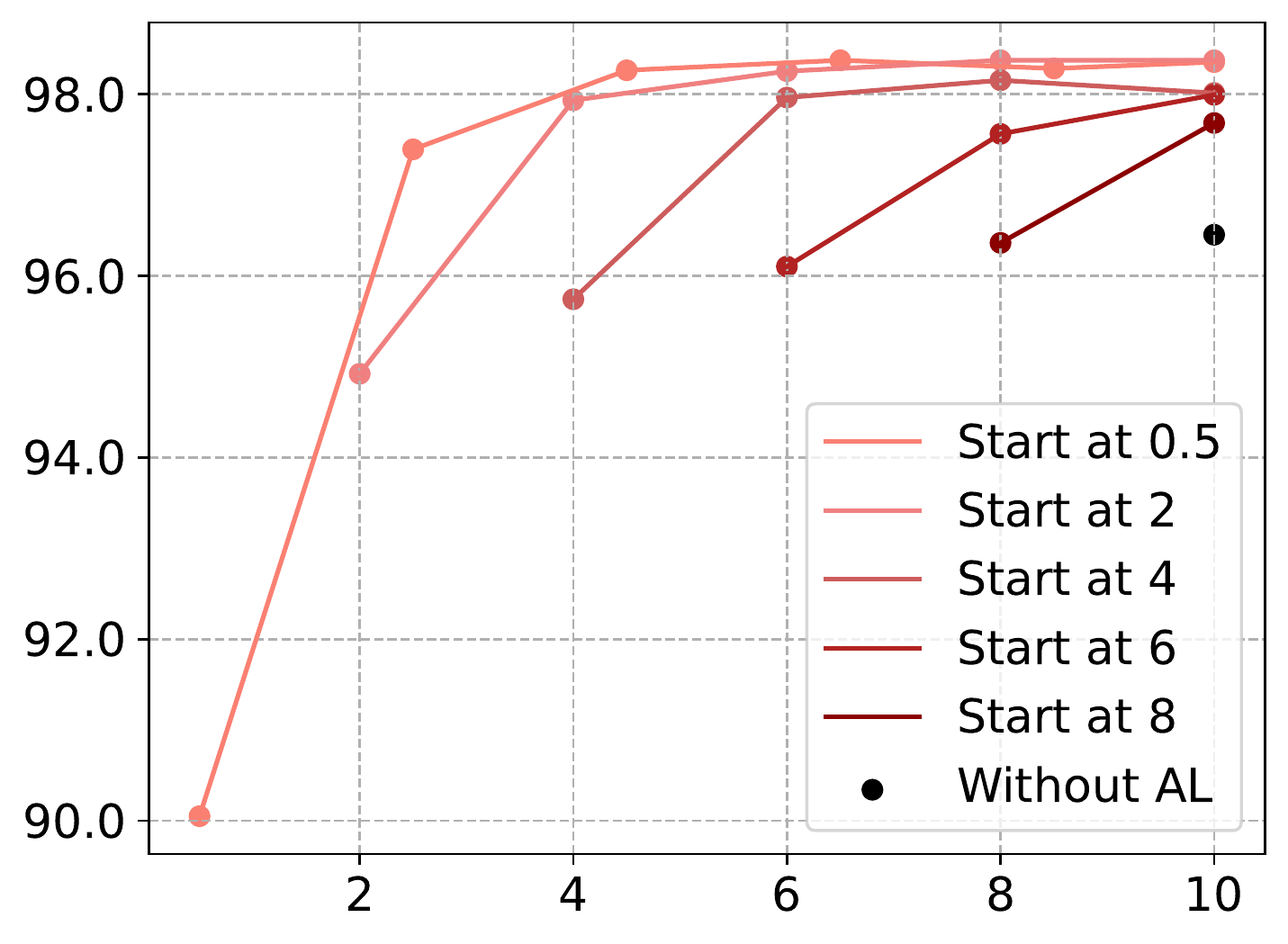}   %以pic.jpg的0.4倍大小输出
		\end{minipage}
	}
	\hspace{1cm}
	\subfigure[CIFAR10] %第一张子图
	{
		\begin{minipage}{5cm}
			\centering          %子图居中
			\includegraphics[scale=0.35]{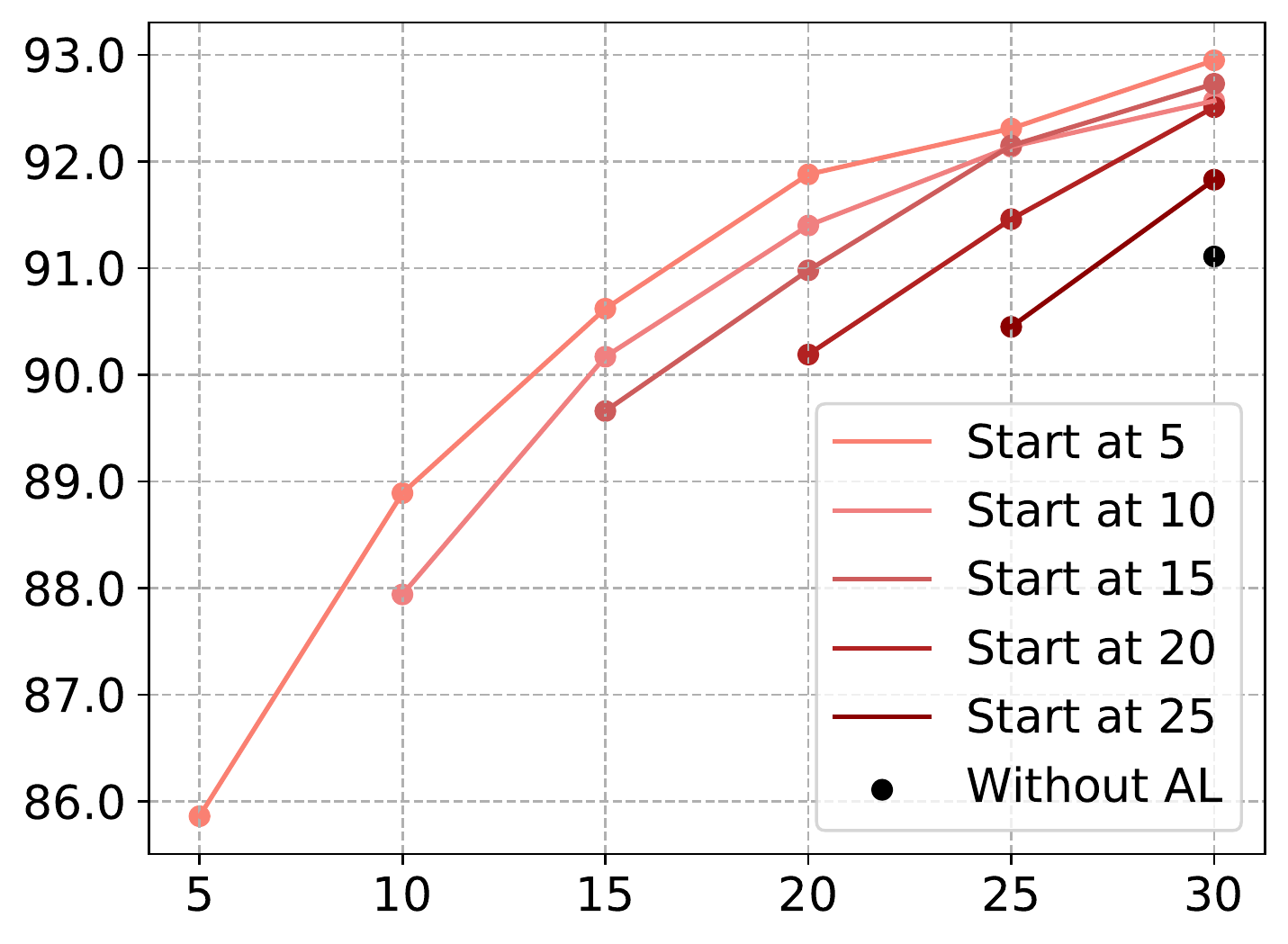}   
		\end{minipage}
	}
	\hspace{0.2cm}
    \vspace{-0.1cm}
 	\caption{The accuracy achieved by SSLDiff2AugDirect when using different initial budgets (\%). The x-axis represents the start budget measured in percentage, and the y-axis stands for the accuracy. We observe that the performance is negatively correlated with the initial budget. This indicates that we should apply the SSAL method as early as possible to achieve better performance.} % 
	\label{fig:ssl_start}  %图片引用标记
% 	\vspace{-0.2cm}
\end{figure*}

\section{In-depth Investigation of SSAL Algorithms}
\label{sec:sslal}
According to Section \ref{sec:empiricalresults}, SSAL is the most promising DAL method as it achieves the best performance across multiple datasets. In this section, we aim to conduct an in-depth analysis of the SSAL method and seek to provide practical guidance to users.

% \subsection{Start SSAL as Early as Possible}
\subsection{{ Start SSAL Early}}

The previous session shows the effectiveness of the SSAL methods. In this part, we investigate when to start the SSAL process would be the best, i.e., we want to decide a proper nStart for a given nEnd. To this end, we fix the total labeling budget and start the SSAL process at different points. Specifically, we set the total budget as 10\% and 30\% for MNIST and CIFAR-10, respectively.
Since the SSAL methods have similar performance results, we randomly choose one, e.g., SSLDiff2AugDirect, to generate the result.
The result is shown in Figure \ref{fig:ssl_start}.
% Assuming the query size is fixed, starting the SSAL iteration too early would introduce more training iteration and hence more cost. 

On the one hand, intuitively, starting the SSAL too late would result in poor performance. The later the SSAL process starts, the more samples are randomly sampled, and the accuracy is closer to SSLRandom. This can be validated by Figure \ref{fig:ssl_start}.
For example, we get around 2 percent accuracy loss if we start from 10\% instead of 2\% on MNIST. This accuracy loss is enormous.
On the other hand, some may argue that starting the SSAL process too early will also cause problems. Because when the labeled data is highly sparse, SSAL can perform worse than SSLRandom \citep{DBLP:journals/corr/abs-1912-05361}.
To consider this, we start the process from 0.5\% for MNIST, and the result shows that the final accuracy achieved by this starting point is the highest. 
% We think the reason lies in the amount of the total budget. 
% As long as the total budget is not extremely small, the final accuracy will be guaranteed if we start early.
% Hence, we suggest starting the SSAL process as early as possible to achieve the best accuracy.
{ Hence, once the randomly selected samples can make the training process go well, we can consider starting the SSAL process.}

 \begin{figure}
\begin{floatrow}
\capbtabbox{%
\resizebox{0.9\linewidth}{!}{
\begin{tabular}{@{}ccc@{}}
\toprule
Dataset       & \begin{tabular}[c]{@{}c@{}}Average Training  \\ Sample Per Class\end{tabular} & \multicolumn{1}{l}{\begin{tabular}[c]{@{}l@{}}Performance\\ Improvement\end{tabular}} \\ \midrule
MNIST         & 6000                                                                & 2.07                                                                                  \\
CIFAR-10      & 5000                                                                & 1.26                                                                                  \\
GTSRB         & 911                                                                 & 1.07                                                                                  \\
Tiny-ImageNet & 500                                                                 & 1.06                                                                                  \\ \bottomrule

\label{table:avg_sample_and_perf}
\end{tabular}
}
}{%
  \caption{The average training sample per class and the performance improvement for different datasets.}
}
\ffigbox{%
%   \begin{figure}
    \centering
    \includegraphics[width=0.85\linewidth]{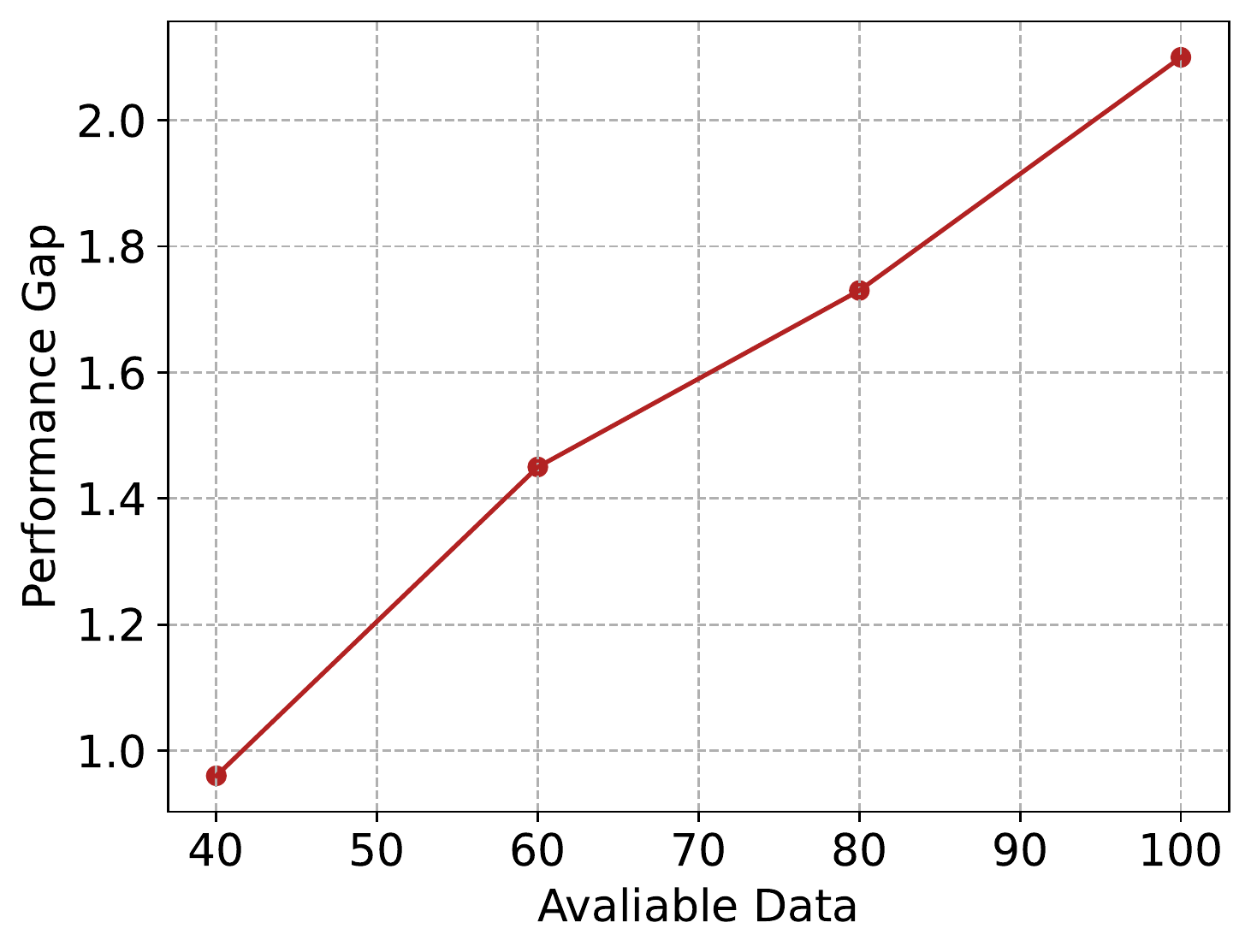}
    \label{fig:available_unlabeled}
% \end{figure}
}{%
  \caption{The performance gap between SSLDiff2AugDirect and SSLRandom w.r.t the number of unlabeled data on CIFAR10 dataset.}
}
\end{floatrow}
\end{figure}
\subsection{The More Unlabeled Data, the Better}
From Table \ref{table:auc}, we observe that the effectiveness of SSAL methods is different for different datasets. 
We use the performance difference between a specific SSAL method and SSLRandom to reflect the effectiveness of that SSAL method. 
For example, the performance of SSLDiff2AugDirect over SSLRandom are 3.49/1.68 = 2.07, 5.78/4.57=1.26, 1.19/1.11=1.07, and 6.33/5.97=1.06 for MNIST, CIFAR-10, GTSRB, and ImageNet, respectively. 
We hypothesize that the performance improvement of SSAL methods is correlated to the number of unlabeled samples available for selection. 
As shown in Table \ref{table:avg_sample_and_perf}, the average number of samples are 6000, 5000, 911, and 500 for MNIST, CIFAR10, GTSRB, and Tiny-ImageNet, respectively. 
More average samples per class mean more data is available for selection, in which case we have a higher probability of obtaining valuable samples.

To validate this hypothesis, we study the effectiveness of SSAL methods by fixing the dataset and only varying the amount of unlabeled data. Specifically, we initialize the model with 5000 labeled images randomly drawn from the CIFAR10 dataset. Then, we provide the model with different amounts of unlabeled data and use SSLDiff2AugDirect for model training and sample selection. The result is shown in Figure \ref{fig:available_unlabeled}. The x-axis of Figure \ref{fig:available_unlabeled} stands for the percentage of data selected from the entire CIFAR10 dataset to train the model. For example, if the selected portion is 40\%, then the amount of unlabeled data equals to 40\% $\times$ 50000 - 5000 = 15000 images. We observe that the performance gap between SSLDiff2AugDirect and SSLRandom increases with the increase of unlabeled data. This result indicates that active data selection performs better if more unlabeled data is available. Hence, we recommend the user collect more unlabeled data before performing SSAL.

\begin{figure}[t]
    \centering
    \includegraphics[width=0.45\linewidth]{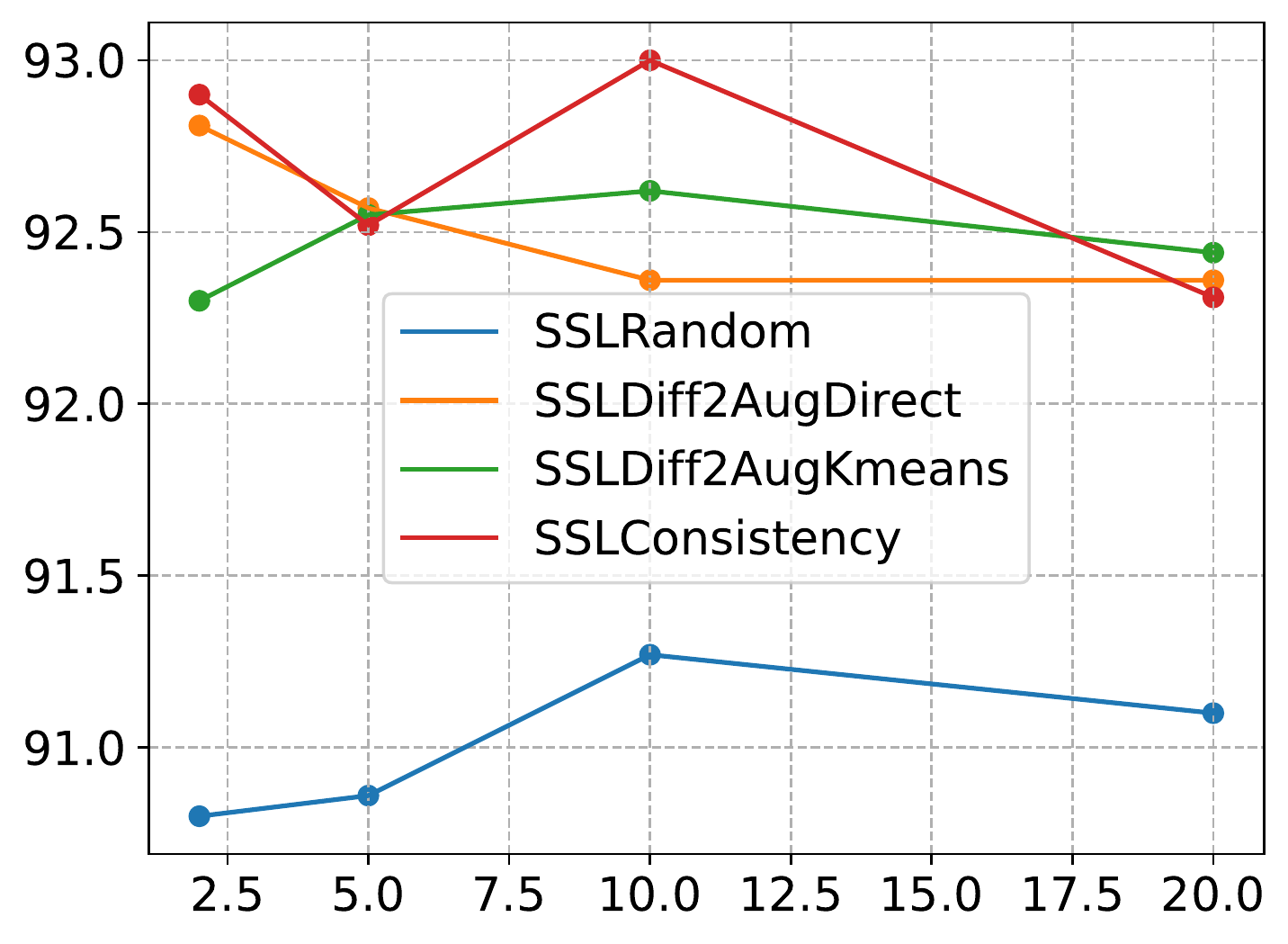}
    \caption{The influence of query size on the performance of SSAL methods.}
    \label{fig:ssl_query_size}
\end{figure}
% \end{wrapfigure}

\subsection{The Query Size has Little Influence on SSAL Performance}
\label{subsec:query_size}

We show the influence of the query size on SSAL methods in Figure \ref{fig:ssl_query_size}. Intuitively, the query size can influence the performance of the SSAL methods. If the query size is too small, the selected samples are too less to be statistically significant. If the query size is too big, the selection would be coarse-grained.
In this experiment, we fix the starting budget as 10\% and the total budget as 50\%. Then, we train the model by varying the query size in each iteration. Despite the performance of SLDiff2AugDirect and SSLConsistency decreasing slightly, their performance is still much higher than SSLRandom. Hence, the SSAL methods can perform well in various query sizes.
% when we increase the query size. This is because the redundancy can increase when we select more samples at once. However, 

% \input{explanation}
\section{Conclusion}
\label{sec:conclusion}

This paper studies the effectiveness of existing deep active learning methods. Our study is the most comprehensive one, which covers \nomethod \ DAL methods, including the state-of-the-art SAL and SSAL ones. Through extensive experiments, we observe that the emerging SSAL techniques provide promising results. Specifically, the traditional SAL methods can hardly beat random selection, and no SAL method can consistently outperform others. However, the SSLAL methods can easily surpass all SAL methods, and we find that active sample selection in SSAL can bring huge performance improvement. In this paper, we also conduct an in-depth analysis of the SSAL methods and give two guidances to the practitioners. First, one should perform SSAL as early as possible and seek more unlabeled data whenever possible to achieve better performance.

\newpage

% \bibliographystyle{plainnat}
% \bibliography{ref.bib}
\bibliographystyle{iclr2023_conference}
\bibliography{iclr2023_conference}

\begin{thebibliography}{30}
\providecommand{\natexlab}[1]{#1}
\providecommand{\url}[1]{\texttt{#1}}
\expandafter\ifx\csname urlstyle\endcsname\relax
  \providecommand{\doi}[1]{doi: #1}\else
  \providecommand{\doi}{doi: \begingroup \urlstyle{rm}\Url}\fi

\bibitem[Arazo et~al.(2019)Arazo, Ortego, Albert, O'Connor, and
  Mcguinness]{2019Pseudo}
E.~Arazo, D.~Ortego, P.~Albert, N.~E. O'Connor, and K.~Mcguinness.
\newblock Pseudo-labeling and confirmation bias in deep semi-supervised
  learning.
\newblock 2019.

\bibitem[Ash et~al.(2020)Ash, Zhang, Krishnamurthy, Langford, and
  Agarwal]{iclrAshZK0A20}
Jordan~T. Ash, Chicheng Zhang, Akshay Krishnamurthy, John Langford, and Alekh
  Agarwal.
\newblock Deep batch active learning by diverse, uncertain gradient lower
  bounds.
\newblock In \emph{8th International Conference on Learning Representations,
  {ICLR} 2020, Addis Ababa, Ethiopia, April 26-30, 2020}. OpenReview.net, 2020.
\newblock URL \url{https://openreview.net/forum?id=ryghZJBKPS}.

\bibitem[Beck et~al.(2021)Beck, Sivasubramanian, Dani, Ramakrishnan, and
  Iyer]{DBLP:journals/corr/abs-2106-15324}
Nathan Beck, Durga Sivasubramanian, Apurva Dani, Ganesh Ramakrishnan, and
  Rishabh~K. Iyer.
\newblock Effective evaluation of deep active learning on image classification
  tasks.
\newblock \emph{CoRR}, abs/2106.15324, 2021.
\newblock URL \url{https://arxiv.org/abs/2106.15324}.

\bibitem[Beluch et~al.(2018)Beluch, Genewein, N{\"u}rnberger, and
  K{\"o}hler]{Beluch2018ThePO}
William~H. Beluch, Tim Genewein, A.~N{\"u}rnberger, and Jan~M. K{\"o}hler.
\newblock The power of ensembles for active learning in image classification.
\newblock \emph{2018 IEEE/CVF Conference on Computer Vision and Pattern
  Recognition}, pp.\  9368--9377, 2018.

\bibitem[Berthelot et~al.(2019)Berthelot, Carlini, Goodfellow, Papernot,
  Oliver, and Raffel]{mixmatch}
David Berthelot, Nicholas Carlini, Ian~J. Goodfellow, Nicolas Papernot, Avital
  Oliver, and Colin Raffel.
\newblock Mixmatch: {A} holistic approach to semi-supervised learning.
\newblock In Hanna~M. Wallach, Hugo Larochelle, Alina Beygelzimer, Florence
  d'Alch{\'{e}}{-}Buc, Emily~B. Fox, and Roman Garnett (eds.), \emph{Advances
  in Neural Information Processing Systems 32: Annual Conference on Neural
  Information Processing Systems 2019, NeurIPS 2019, December 8-14, 2019,
  Vancouver, BC, Canada}, pp.\  5050--5060, 2019.
\newblock URL
  \url{https://proceedings.neurips.cc/paper/2019/hash/1cd138d0499a68f4bb72bee04bbec2d7-Abstract.html}.

\bibitem[Caramalau et~al.(2021)Caramalau, Bhattarai, and
  Kim]{DBLP:conf/cvpr/CaramalauBK21}
Razvan Caramalau, Binod Bhattarai, and Tae{-}Kyun Kim.
\newblock Sequential graph convolutional network for active learning.
\newblock In \emph{{IEEE} Conference on Computer Vision and Pattern
  Recognition, {CVPR} 2021, virtual, June 19-25, 2021}, pp.\  9583--9592.
  Computer Vision Foundation / {IEEE}, 2021.
\newblock URL
  \url{https://openaccess.thecvf.com/content/CVPR2021/html/Caramalau\_Sequential\_Graph\_Convolutional\_Network\_for\_Active\_Learning\_CVPR\_2021\_paper.html}.

\bibitem[Chan et~al.(2021)Chan, Li, and Oymak]{Chan2021OnTM}
Yao-Chun Chan, Mingchen Li, and Samet Oymak.
\newblock On the marginal benefit of active learning: Does self-supervision eat
  its cake?
\newblock \emph{ICASSP 2021 - 2021 IEEE International Conference on Acoustics,
  Speech and Signal Processing (ICASSP)}, pp.\  3455--3459, 2021.

\bibitem[Christoph~M.(2020)]{DBLP:conf/wacv/0007T20}
Radu~Timofte Christoph~M.
\newblock Adversarial sampling for active learning.
\newblock In \emph{{IEEE} Winter Conference on Applications of Computer Vision,
  {WACV} 2020, Snowmass Village, CO, USA, March 1-5, 2020}, pp.\  3060--3068.
  {IEEE}, 2020.
\newblock \doi{10.1109/WACV45572.2020.9093556}.
\newblock URL \url{https://doi.org/10.1109/WACV45572.2020.9093556}.

\bibitem[Citovsky et~al.(2021)Citovsky, DeSalvo, Gentile, Karydas, Rajagopalan,
  Rostamizadeh, and Kumar]{citovsky2021batch}
Gui Citovsky, Giulia DeSalvo, Claudio Gentile, Lazaros Karydas, Anand
  Rajagopalan, Afshin Rostamizadeh, and Sanjiv Kumar.
\newblock Batch active learning at scale.
\newblock In A.~Beygelzimer, Y.~Dauphin, P.~Liang, and J.~Wortman Vaughan
  (eds.), \emph{Advances in Neural Information Processing Systems}, 2021.
\newblock URL \url{https://openreview.net/forum?id=zzdf0CirJM4}.

\bibitem[Ducoffe~M.(2018)]{ducoffe2018adversarial}
Precioso~F. Ducoffe~M.
\newblock Adversarial active learning for deep networks: a margin based
  approach.
\newblock \emph{arXiv preprint arXiv:1802.09841}, 2018.

\bibitem[Gal et~al.(2017{\natexlab{a}})Gal, Islam, and
  Ghahramani]{DBLP:conf/icml/GalIG17}
Yarin Gal, Riashat Islam, and Zoubin Ghahramani.
\newblock Deep bayesian active learning with image data.
\newblock In Doina Precup and Yee~Whye Teh (eds.), \emph{Proceedings of the
  34th International Conference on Machine Learning, {ICML} 2017, Sydney, NSW,
  Australia, 6-11 August 2017}, volume~70, pp.\  1183--1192. {PMLR},
  2017{\natexlab{a}}.
\newblock URL \url{http://proceedings.mlr.press/v70/gal17a.html}.

\bibitem[Gal et~al.(2017{\natexlab{b}})Gal, Islam, and Ghahramani]{gal2017deep}
Yarin Gal, Riashat Islam, and Zoubin Ghahramani.
\newblock Deep bayesian active learning with image data.
\newblock In \emph{International Conference on Machine Learning}, pp.\
  1183--1192. PMLR, 2017{\natexlab{b}}.

\bibitem[Gao et~al.(2020)Gao, Zhang, Yu, Arik, Davis, and Pfister]{Consistency}
Mingfei Gao, Zizhao Zhang, Guo Yu, Sercan~{\"{O}}mer Arik, Larry~S. Davis, and
  Tomas Pfister.
\newblock Consistency-based semi-supervised active learning: Towards minimizing
  labeling cost.
\newblock In Andrea Vedaldi, Horst Bischof, Thomas Brox, and Jan{-}Michael
  Frahm (eds.), \emph{Computer Vision - {ECCV} 2020 - 16th European Conference,
  Glasgow, UK, August 23-28, 2020, Proceedings, Part {X}}, volume 12355 of
  \emph{Lecture Notes in Computer Science}, pp.\  510--526. Springer, 2020.
\newblock \doi{10.1007/978-3-030-58607-2\_30}.
\newblock URL \url{https://doi.org/10.1007/978-3-030-58607-2\_30}.

\bibitem[Hoi et~al.(2006)Hoi, Jin, Zhu, and Lyu]{HoiJZL06}
Steven C.~H. Hoi, Rong Jin, Jianke Zhu, and Michael~R. Lyu.
\newblock Batch mode active learning and its application to medical image
  classification.
\newblock In \emph{Machine Learning, Proceedings of the Twenty-Third
  International Conference {(ICML} 2006), Pittsburgh, Pennsylvania, USA, June
  25-29, 2006}, volume 148, pp.\  417--424. {ACM}, 2006.
\newblock \doi{10.1145/1143844.1143897}.
\newblock URL \url{https://doi.org/10.1145/1143844.1143897}.

\bibitem[Houben et~al.(2013)Houben, Stallkamp, Salmen, Schlipsing, and
  Igel]{Houben-IJCNN-2013}
Sebastian Houben, Johannes Stallkamp, Jan Salmen, Marc Schlipsing, and
  Christian Igel.
\newblock Detection of traffic signs in real-world images: The {G}erman
  {T}raffic {S}ign {D}etection {B}enchmark.
\newblock In \emph{International Joint Conference on Neural Networks}, number
  1288, 2013.

\bibitem[Hu et~al.(2021)Hu, Guo, Cordy, Xie, Ma, Papadakis, and
  Traon]{DBLP:conf/kbse/Hu0CXMPT21}
Qiang Hu, Yuejun Guo, Maxime Cordy, Xiaofei Xie, Wei Ma, Mike Papadakis, and
  Yves~Le Traon.
\newblock Towards exploring the limitations of active learning: An empirical
  study.
\newblock In \emph{36th {IEEE/ACM} International Conference on Automated
  Software Engineering, {ASE} 2021, Melbourne, Australia, November 15-19,
  2021}, pp.\  917--929. {IEEE}, 2021.
\newblock \doi{10.1109/ASE51524.2021.9678672}.
\newblock URL \url{https://doi.org/10.1109/ASE51524.2021.9678672}.

\bibitem[Krizhevsky et~al.(2009)Krizhevsky, Hinton,
  et~al.]{krizhevsky2009learning}
Alex Krizhevsky, Geoffrey Hinton, et~al.
\newblock Learning multiple layers of features from tiny images.
\newblock Technical report, Citeseer, 2009.

\bibitem[Le \& Yang(2015)Le and Yang]{Le2015TinyIV}
Ya~Le and Xuan~S. Yang.
\newblock Tiny imagenet visual recognition challenge.
\newblock 2015.

\bibitem[LeCun et~al.(1998)LeCun, Bottou, Bengio, and
  Haffner]{lecun1998gradient}
Yann LeCun, L{\'e}on Bottou, Yoshua Bengio, and Patrick Haffner.
\newblock Gradient-based learning applied to document recognition.
\newblock \emph{Proceedings of the IEEE}, 86\penalty0 (11):\penalty0
  2278--2324, 1998.

\bibitem[Mittal et~al.(2019)Mittal, Tatarchenko, {\c{C}}i{\c{c}}ek, and
  Brox]{DBLP:journals/corr/abs-1912-05361}
Sudhanshu Mittal, Maxim Tatarchenko, {\"{O}}zg{\"{u}}n {\c{C}}i{\c{c}}ek, and
  Thomas Brox.
\newblock Parting with illusions about deep active learning.
\newblock \emph{CoRR}, abs/1912.05361, 2019.
\newblock URL \url{http://arxiv.org/abs/1912.05361}.

\bibitem[Munjal et~al.(2020)Munjal, Hayat, Hayat, Sourati, and
  Khan]{DBLP:journals/corr/abs-2002-09564}
Prateek Munjal, Nasir Hayat, Munawar Hayat, Jamshid Sourati, and Shadab Khan.
\newblock Towards robust and reproducible active learning using neural
  networks.
\newblock \emph{CoRR}, abs/2002.09564, 2020.
\newblock URL \url{https://arxiv.org/abs/2002.09564}.

\bibitem[Nissim et~al.(2014)Nissim, Cohen, Moskovitch, Shabtai, Edry, Bar{-}Ad,
  and Elovici]{NissimCMSEBE14}
Nir Nissim, Aviad Cohen, Robert Moskovitch, Asaf Shabtai, Mattan Edry, Oren
  Bar{-}Ad, and Yuval Elovici.
\newblock {ALPD:} active learning framework for enhancing the detection of
  malicious {PDF} files.
\newblock In \emph{{IEEE} Joint Intelligence and Security Informatics
  Conference, {JISIC} 2014, The Hague, The Netherlands, 24-26 September, 2014},
  pp.\  91--98. {IEEE}, 2014.
\newblock \doi{10.1109/JISIC.2014.23}.
\newblock URL \url{https://doi.org/10.1109/JISIC.2014.23}.

\bibitem[Sener \& Savarese(2018)Sener and Savarese]{iclrSenerS18}
Ozan Sener and Silvio Savarese.
\newblock Active learning for convolutional neural networks: {A} core-set
  approach.
\newblock In \emph{6th International Conference on Learning Representations,
  {ICLR} 2018, Vancouver, BC, Canada, April 30 - May 3, 2018, Conference Track
  Proceedings}. OpenReview.net, 2018.
\newblock URL \url{https://openreview.net/forum?id=H1aIuk-RW}.

\bibitem[Shui et~al.(2020)Shui, Zhou, Gagn{\'e}, and Wang]{shui2020deep}
Changjian Shui, Fan Zhou, Christian Gagn{\'e}, and Boyu Wang.
\newblock Deep active learning: Unified and principled method for query and
  training.
\newblock In \emph{International Conference on Artificial Intelligence and
  Statistics}, pp.\  1308--1318. PMLR, 2020.

\bibitem[Sinha et~al.(2019)Sinha, Ebrahimi, and Darrell]{iccvSinhaED19}
Samarth Sinha, Sayna Ebrahimi, and Trevor Darrell.
\newblock Variational adversarial active learning.
\newblock In \emph{2019 {IEEE/CVF} International Conference on Computer Vision,
  {ICCV} 2019, Seoul, Korea (South), October 27 - November 2, 2019}, pp.\
  5971--5980. {IEEE}, 2019.
\newblock \doi{10.1109/ICCV.2019.00607}.
\newblock URL \url{https://doi.org/10.1109/ICCV.2019.00607}.

\bibitem[Song et~al.(2019)Song, Berthelot, and Rostamizadeh]{MMA}
Shuang Song, David Berthelot, and Afshin Rostamizadeh.
\newblock Combining mixmatch and active learning for better accuracy with fewer
  labels.
\newblock \emph{CoRR}, abs/1912.00594, 2019.
\newblock URL \url{http://arxiv.org/abs/1912.00594}.

\bibitem[Wang \& Shang(2014)Wang and Shang]{DBLP:conf/ijcnn/WangS14}
Dan Wang and Yi~Shang.
\newblock A new active labeling method for deep learning.
\newblock In \emph{2014 International Joint Conference on Neural Networks,
  {IJCNN} 2014, Beijing, China, July 6-11, 2014}, pp.\  112--119. {IEEE}, 2014.
\newblock \doi{10.1109/IJCNN.2014.6889457}.
\newblock URL \url{https://doi.org/10.1109/IJCNN.2014.6889457}.

\bibitem[Yoo \& Kweon(2019)Yoo and Kweon]{DBLP:conf/cvpr/YooK19}
Donggeun Yoo and In~So Kweon.
\newblock Learning loss for active learning.
\newblock In \emph{{IEEE} Conference on Computer Vision and Pattern
  Recognition, {CVPR} 2019, Long Beach, CA, USA, June 16-20, 2019}, pp.\
  93--102. Computer Vision Foundation / {IEEE}, 2019.
\newblock \doi{10.1109/CVPR.2019.00018}.
\newblock URL
  \url{http://openaccess.thecvf.com/content\_CVPR\_2019/html/Yoo\_Learning\_Loss\_for\_Active\_Learning\_CVPR\_2019\_paper.html}.

\bibitem[Yuan et~al.(2019)Yuan, Hou, Xiao, Cao, Guan, and
  Nie]{DBLP:journals/kbs/YuanHXCGN19}
Jin Yuan, Xingxing Hou, Yaoqiang Xiao, Da~Cao, Weili Guan, and Liqiang Nie.
\newblock Multi-criteria active deep learning for image classification.
\newblock \emph{Knowl. Based Syst.}, 172:\penalty0 86--94, 2019.
\newblock \doi{10.1016/j.knosys.2019.02.013}.
\newblock URL \url{https://doi.org/10.1016/j.knosys.2019.02.013}.

\bibitem[Zhdanov(2019)]{Zhdanov2019DiverseMA}
Fedor Zhdanov.
\newblock Diverse mini-batch active learning.
\newblock \emph{ArXiv}, abs/1901.05954, 2019.

\end{thebibliography}

\newpage
\appendix
\section{Appendix}
\subsection{Experimental Settings of different DAL paper}
We summarize the detailed settings of some representative DAL papers in Table \ref{table:hyperparameters}. As we can observe, these hyperparameters differ a lot in different papers, including the start budget size (Start), step size, learning rate (LR), training epoch, accuracy calculation criteria, and optimizer. This is the reason why we cannot compare their performance only by looking for the reported numbers in their paper. Our empirical study unifies these settings and provides a fair comparison of these methods.

\begin{table}[H]
\centering
\caption{Hyperparameters used by different active learning methods on CIFAR-10.}
\label{table:hyperparameters}
\resizebox{\linewidth}{!}{  
 \begin{tabular}{ccccccc}
\toprule
\textbf{Method}       & \textbf{Start} & \textbf{Step Size} & \textbf{LR} & \textbf{Training Epoch}                                                         & \textbf{Accuracy Calculation}              & \textbf{Optimizer} \\ \midrule
BadgeSampling         & 0.2\%          & 0.2\%/2\%/20\%     & 1e-3        & \begin{tabular}[c]{@{}c@{}}Terminate when 99\% \\ accuracy archived\end{tabular} & After train                    & Adam               \\ 
ClusterMargin & 10\%           & 10\%               & 1e-3        & -                                                                               & -                              & batch SGD          \\ 
CoreSet               & 10\%           & 10\%               & 1e-3        & -                                                                               &                                & RMSProp            \\ 
KMeansSampling        & 10\%           & 10\%               & 1e-3        & -                                                                               &                                & RMSProp            \\ 
LearningLoss          & 2\%            & 2\%                & 1e-1, 1e-2  & 200                                                                             & After train                    & SGD                \\ 
MCADL                 & 4\%            & 128                & -           & -                                                                               & -                              & Adam               \\ 
WAAL                  & 2\%            & 2\%                & 1e-2        & 80                                                                              & After train                    & SGD                \\ 
CoreGCN/UncertainGCN  & 2\%            & 2\%                & 1e-1, 1e-2  & 200                                                                             & -                              & SGD                \\ 
Ensemble              & 0.4\%          & 0.8\%              & -           & 150                                                                             & -                              & RMSprop            \\ 
VAAL                  & 10\%           & 5\%                & 1e-2        & 100                                                                             & Use best model on validate set & SGD                \\ \bottomrule
\end{tabular}
}

\end{table}

\subsection{Detailed performance of DAL methods}
\begin{figure}[H]
\centering
\includegraphics[width=\linewidth]{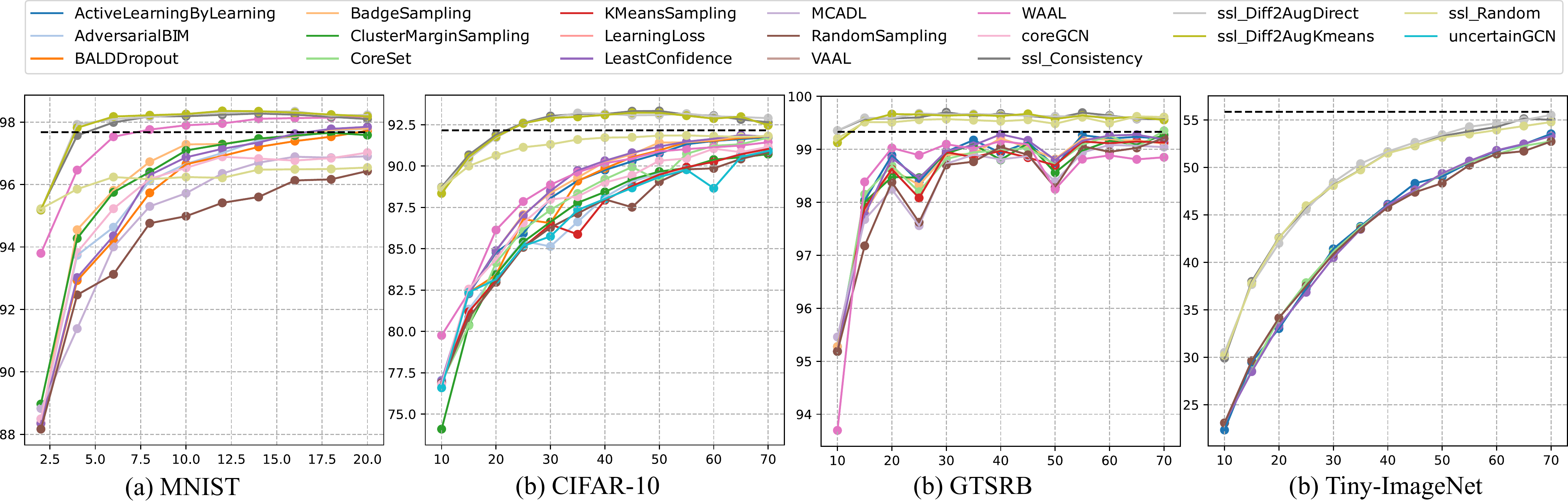}
\caption{The performance of the deep active learning techniques. Note that for better visualization, we only draw the lines for methods that is better than random selection. The x-axis denotes the budget in percentage, and the y-axis denotes the accuracy. }
\label{fig:performance}
\end{figure}

We show the detailed accuracy against the budget curve in Figure \ref{fig:performance}. Note that we also include the performance of Tiny-ImageNet to privide more information. 
Tiny-ImageNet is a subset of the ILSVRC- 2012 classification dataset. It consists of 200 object classes. All images were down-sampled to 64 × 64 × 3 pixels. Tiny-ImageNet is a more practical and complex dataset.
As we can observe, the SSAL methods consistently outperform SAL methods in all datasets.

\subsection{Robustness study of SAL methods}
% \textbf{Robustness.} 
We investigate two factors that can have a large influence on the performance of the SAL method: the initial budget and the query size. If the initial budget is too small, the initial model can be poorly trained. Selecting new data using this model can cause less optimal results. If the initial budget is too large, most of the labeled samples are randomly selected, and hence the model accuracy would be similar to that of random selection. If the query size is too small, the selected samples are too less to be statistically significant. If the query size is too big, the selection would be coarse-grained. To study these two factors, we choose five representative methods: BadgeSampling, LeastConfidence, CoreGCN, ClusterMargin, and BALDDropout. The experiment in this section is performed on CIFAR-10.

\begin{figure*}[]	
    \hspace{0.8cm}
	\subfigure[Different Initial Budget (\%)] %第一张子图
	{
		\begin{minipage}{5cm}
			\centering          %子图居中
			\includegraphics[scale=0.35]{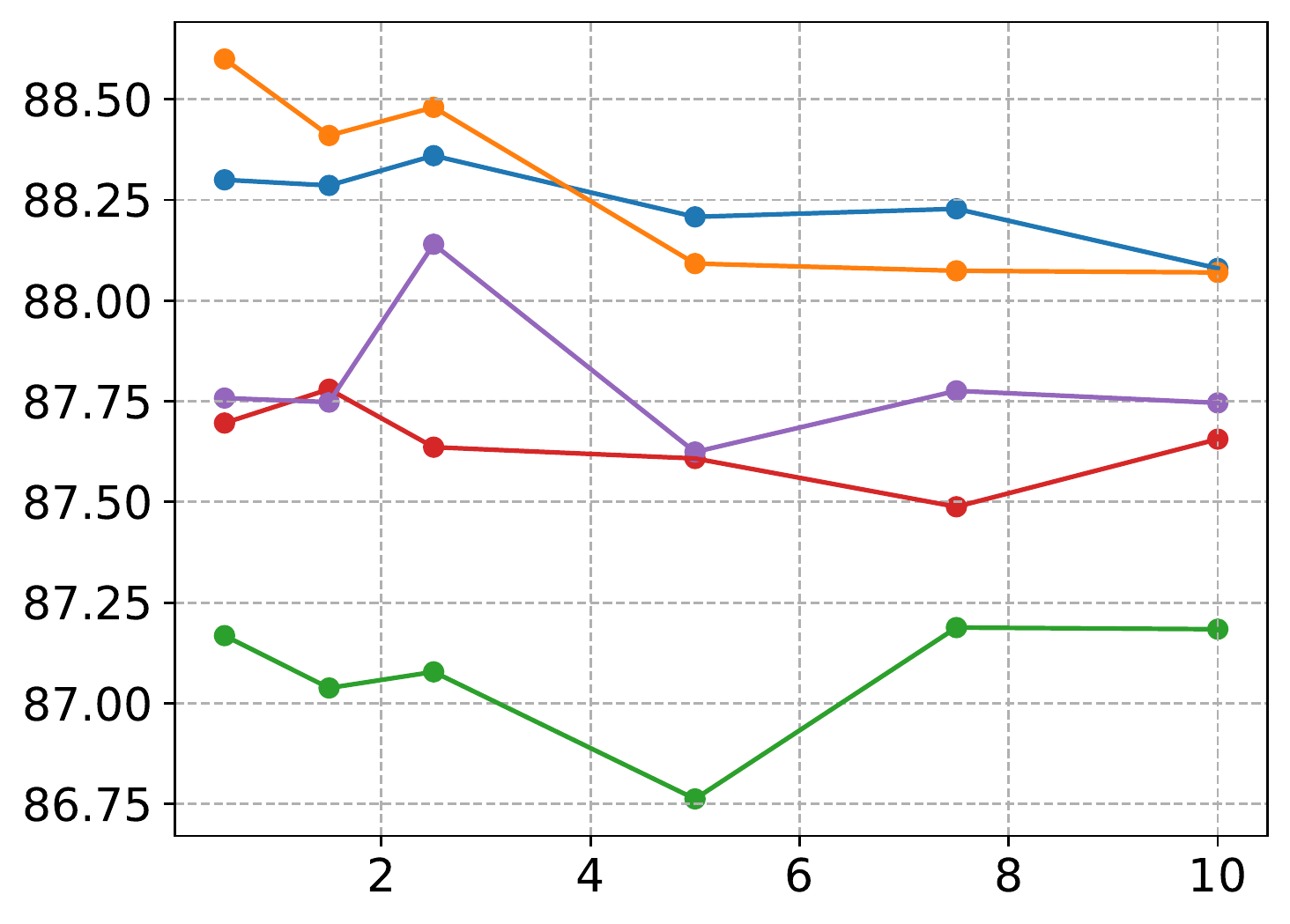}   %以pic.jpg的0.4倍大小输出
		\end{minipage}
	}
	\hspace{1cm}
	\subfigure[Different Query Budget (\%)] %第一张子图
	{
		\begin{minipage}{5cm}
			\centering          %子图居中
			\includegraphics[scale=0.35]{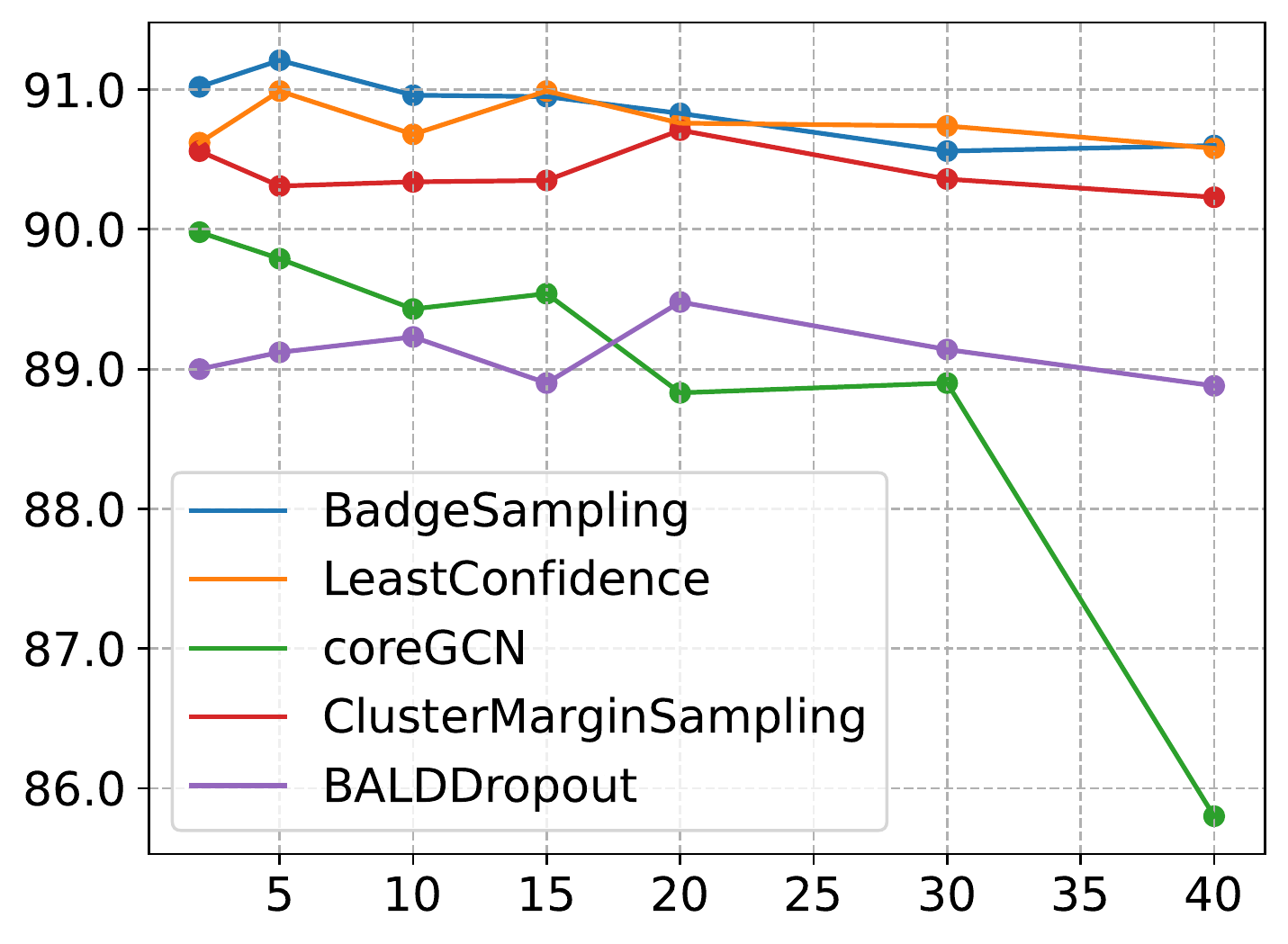}   
		\end{minipage}
	}
	\hspace{0.2cm}
    \vspace{-0.1cm}
	\caption{The performance of five SAL methods under different initial budget and query budget. (a) the impact of the initial budget. The nQuery and nEnd are 5\% and 40\%, respectively. (b) the impact of the query budget. The nStart and nEnd are 10\% and 50\%, respectively.} % 
	\label{fig:robustness}  %图片引用标记
% 	\vspace{-0.2cm}
\end{figure*}

\textbf{Start budget.} The influence of the initial budget on the performance is shown in Figure \ref{fig:robustness} (a). In this experiment, we fix the total budget as 30\% and the query size as 2\% for each method, and train the model with different initial budgets. From Figure \ref{fig:robustness} (a), we observe that the performance of each method does not change much with different initial budgets. 

\textbf{Query size.} The influence of the query size in each DAL iteration is shown in Figure \ref{fig:robustness} (b). In this experiment, we fix the initial budget as 10\% and the end budget as 50\%. Then, we train the model by varying the query size in each iteration. We observe that the performance decreases when we increase the query size. This is because the redundancy can increase when we select more samples at once. Note that our observation is in contrast with \cite{DBLP:journals/corr/abs-2106-15324}, where the authors claims that the query size has little influence on the performance. We suspect the difference is that the query size in \cite{DBLP:journals/corr/abs-2106-15324} is too small to reflect the trend.

\end{document}